\pdfoutput=1

\documentclass[11pt]{article}

\usepackage{acl}
\usepackage{amsmath}
\usepackage{graphicx}
\usepackage{hyperref}
\usepackage{adjustbox} 
\usepackage{times}
\usepackage{latexsym}
\usepackage{booktabs}
\usepackage{multirow} 
\usepackage[T1]{fontenc}

\usepackage[utf8]{inputenc}
\usepackage{bigstrut} 
\usepackage{pdfpages}
\usepackage{microtype}
\usepackage{tabularx} 
\usepackage{inconsolata}

\usepackage{graphicx}

\title{Exploring the Multilingual NLG Evaluation Abilities of LLM-Based Evaluators}

\author{First Author \\
  Affiliation / Address line 1 \\
  Affiliation / Address line 2 \\
  Affiliation / Address line 3 \\
  \texttt{email@domain} \\\And
  Second Author \\
  Affiliation / Address line 1 \\
  Affiliation / Address line 2 \\
  Affiliation / Address line 3 \\
  \texttt{email@domain} \\}

\author{
 \textbf{Jiayi Chang\textsuperscript{1,2}},
 \textbf{Mingqi Gao\textsuperscript{1}},
 \textbf{Xinyu Hu\textsuperscript{1}},
 \textbf{Xiaojun Wan\textsuperscript{1}}
\\
 \textsuperscript{1}Wangxuan Institute of Computer Technology, Peking University \\
 \textsuperscript{2}School of Data Science and Intelligent Media, Communication University of China
\\
 }

\begin{document}
\maketitle
\begin{abstract}
Previous research has shown that LLMs have potential in multilingual NLG evaluation tasks. However, existing research has not fully explored the differences in the evaluation capabilities of LLMs across different languages. To this end, this study provides a comprehensive analysis of the multilingual evaluation performance of 10 recent LLMs, spanning high-resource and low-resource languages through correlation analysis, perturbation attacks, and fine-tuning. We found that 1) excluding the reference answer from the prompt and using large-parameter LLM-based evaluators leads to better performance across various languages; 2) most LLM-based evaluators show a higher correlation with human judgments in high-resource languages than in low-resource languages; 3) in the languages where they are most sensitive to such attacks, they also tend to exhibit the highest correlation with human judgments; and 4) fine-tuning with data from a particular language yields a broadly consistent enhancement in the model’s evaluation performance across diverse languages. Our findings highlight the imbalance in LLMs’ evaluation capabilities across different languages and suggest that low-resource language scenarios deserve more attention.
\end{abstract}

\section{Introduction}

With the emergence of sophisticated large language models (LLMs), evaluators driven by LLMs have gained substantial prominence across a diverse array of natural language generation (NLG) tasks, particularly in the multilingual evaluation domain \citep{doddapaneni2024cross,hada2023large,chollampatt2025cross,lai2023chatgpt}. Conventional evaluation metrics frequently exhibit limitations in their capacity to discern the intricate and nuanced characteristics of NLG \citep{papinesi2002bleu,kryscinski2019evaluating,lin2004rouge,sellam2020bleurt,rei2020comet}, while human evaluation remains infeasible due to its prohibitive cost. Emerging research highlights the considerable promise of LLMs as advanced, automated evaluators \citep{liu2023g,mendoncca2023simple,zheng2023judging,li2023alpacaeval,gao2023human}.

However, existing research has not fully explored the differences in the evaluation capabilities of LLMs across different languages. Therefore, this paper explores this issue. Before investigating the differences in LLMs' evaluation capabilities across languages, we first aim to identify universal principles for selecting and utilizing LLMs for multilingual evaluation (\textbf{RQ1}). We found that large-parameter LLM-based evaluators show a higher correlation with human judgments than small-parameter evaluators, which aligns with expectations. Surprisingly, we also found that \textbf{including a reference answer in the prompt often harms the performance of LLM-based evaluators}, casting doubt on the practice in existing studies of always including a reference answer \citep{doddapaneni2024cross}. These two findings apply across almost every model family and language.

Next, we are interested in whether there are differences in the correlation between LLM-based evaluators and human judgments across languages (\textbf{RQ2}). We discovered that \textbf{most LLM-based evaluators show a higher correlation with human judgments in high-resource languages (English, French, German) than in low-resource languages (Bengali, Hindi, Telugu, Urdu)}. This finding highlights the imbalance in LLMs' evaluation capabilities across different languages.

To further investigate the behavior of LLM-based evaluators in different languages, we designed multilingual perturbation attacks to examine whether LLM-based evaluators show varying sensitivity to such attacks across languages (\textbf{RQ3}). We found that most LLM-based evaluators are more sensitive to perturbation attacks in high-resource languages, and \textbf{in the languages where they are most sensitive to such attacks, they also tend to exhibit the highest correlation with human judgments}. This corroborates the findings from RQ2.

The experimental results of the above RQs demonstrate the limitations of LLM-based evaluators on low-resource languages. After examining the pretraining data of these LLMs, we hypothesize that the lack of training data for low-resource languages contributes to the observed phenomenon. To address this, we aim to explore how training data from different languages affects the multilingual evaluation capabilities of LLMs (\textbf{RQ4}). We found that fine-tuning the model with data in a specific language generally improves its evaluation ability across all languages. However, the model fine-tuned with data in a specific language does not necessarily perform better in that language compared to models fine-tuned with data in other languages.



In summary, our contributions are as follows:
\begin{itemize}
    \item This study provides a comprehensive analysis of the multilingual evaluation capabilities of 10 recent LLMs, spanning high-resource and low-resource languages.

        \item 
        Our findings highlight the imbalance in LLMs' evaluation capabilities across different languages and suggest that low-resource language scenarios deserve more attention.
    \item We curate a high-quality perturbation dataset, encompassing tasks such as title generation and summarization, with rigorous alignment across multiple languages. The associated resources will be made publicly available to facilitate future research.
\end{itemize}

\begin{figure}[h!]
    \centering
    \includegraphics[width=\columnwidth]{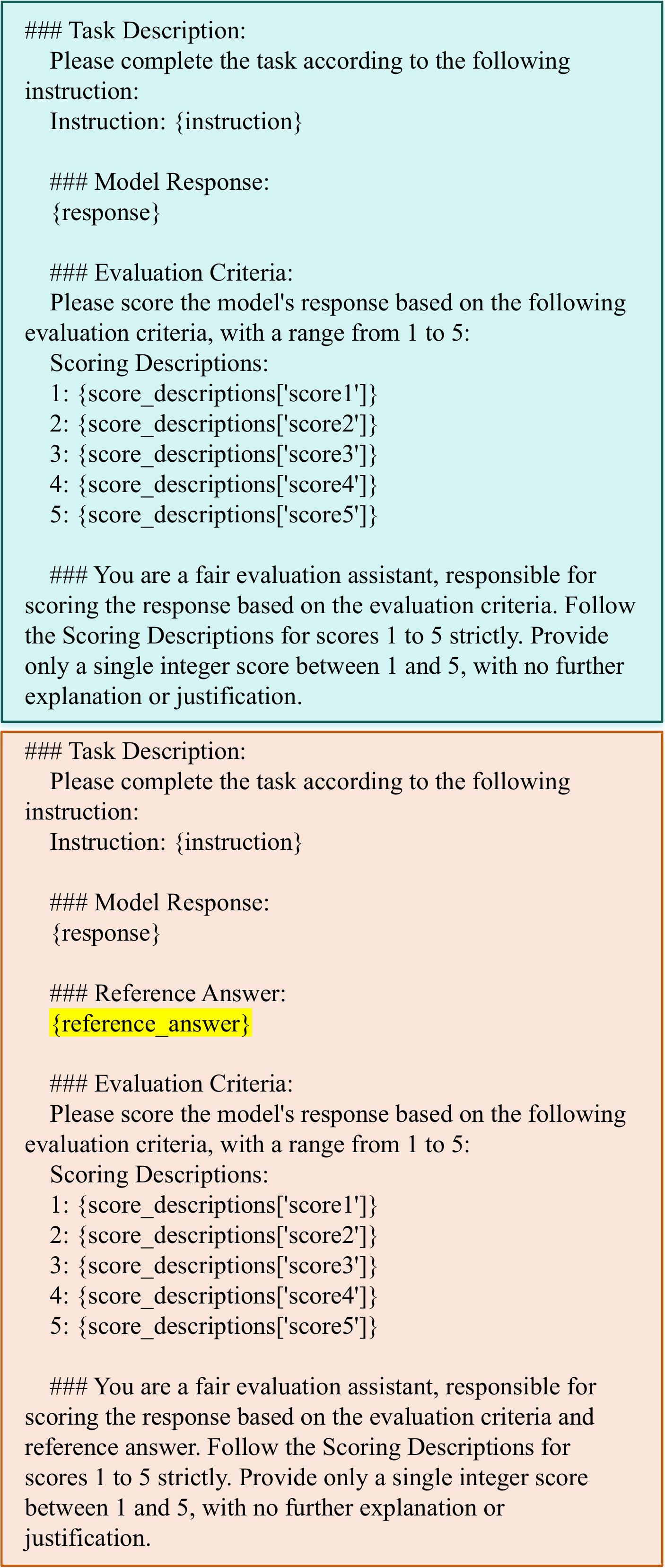} 

    \caption{The instruction for prompting LLMs to evaluate responses based on predefined evaluation criteria without reference answer, contrasted with the approach that incorporates both evaluation criteria and reference answer.}

    \label{fig:1}
\end{figure}

\begin{table*}[h]
\centering
\small
\renewcommand{\arraystretch}{1.2} 
\setlength{\tabcolsep}{6pt} 
\begin{tabular}{l p{10cm}}
\toprule
\multicolumn{1}{c}{\textbf{Disturbance Type}} & \multicolumn{1}{c}{\textbf{Requirement}} \\  
\midrule
Simplicity: Uncommon Phrase  & Replace a phrase or word in the original sentences with an uncommon phrase or word. \\  
Simplicity: Complex Sentence & Convert the original sentences to passive voice or break it into several short sentences to make it more complex. \\  
Non-hallucination: Complement & Insert a clause in the middle of the original sentences or add an adjective before a noun. \\  
Non-hallucination: Continuation & Add a sentence at the end of the original sentences and ensure the relationship between the two sentences. \\  
Non-contradiction: Different Entity & For each sentence in the original sentences, replace one word with another word of the same category but with a completely different meaning. \\  
Non-contradiction: Negation & Change part of the original sentences from affirmation to negation or from negation to affirmation. \\  
\bottomrule
\end{tabular}
\caption{Disturbance types and their corresponding requirements for title generation task.}
\label{tab:disturbance}
\end{table*}

\section{Research Design}

\subsection{LLMs, Language Selection and Datasets}
\textbf{LLMs}
We select ten recent LLMs to investigate the factors that influence their multilingual evaluation capabilities, including the Qwen family, the GLM family, the Llama family, and other models. Specifically, they are Qwen2.5-7B-Instruct \citep{yang2024qwen2}, Qwen2.5-72B-Instruct \citep{yang2024qwen2}, Qwen-turbo, glm-4-flash \citep{glm2024chatglm}, glm-4-9B-chat, Llama-3.1-8B-Instruct \citep{dubey2024llama}, Llama-3.1-70B-Instruct \citep{dubey2024llama}, Gemini-1.5pro \citep{team2024gemini}, ChatGPT-3.5-turbo, and Gemma-2-9B-it \citep{team2024gemma}. 

\textbf{Languages}
We curated a diverse set of languages encompassing high-resource and low-resource categories. Among the high-resource languages are English, German, and French \citep{nigatu2024zeno,babu2021xls}. Additionally, Bengali, Hindi, Telugu, and Urdu exemplify the low-resource languages in our selection \citep{nigatu2024zeno}.

\textbf{Datasets}
We employ the RECON dataset \citep{doddapaneni2024cross}, which serves as the foundation for our investigations into RQ1, RQ2, and RQ4. This dataset spans aforementioned seven languages, comprising 500 items per language, amounting to a total of 3,500. Notably, the multilingual data is meticulously aligned, ensuring that evaluation scores for identical texts remain consistent across all languages. Additionally, in order to conduct perturbation attacks, we complement and expand the MTG dataset \citep{chen2021mtg}, which is limited to the high-resource languages we mentioned, into a new dataset, the Multilingual Perturbation Evaluation (MPE), to bridge this gap. The new dataset encompasses 20,000 samples, with specific details on dataset construction provided in \autoref{sec:The Sensitivity of LLMs to Perturbed Texts}.

\subsection{Correlation Analysis}

We analyze the correlation between their evaluation scores and human judgments to measure their evaluation capabilities. Our investigation primarily centers on whether the prompts contain reference answers, the model's parameter scale, and whether different languages influence the evaluation capabilities of LLMs. We treat the first two factors as language-independent variables for investigation, corresponding to RQ1, which addresses universal principles for better selecting and utilizing LLMs for multilingual evaluation. The impact of different languages corresponds to RQ2, which examines whether there are differences in the correlation between LLMs and human evaluations across languages.
\begin{figure*}[h] 
    \centering 
    \includegraphics[width=\textwidth]{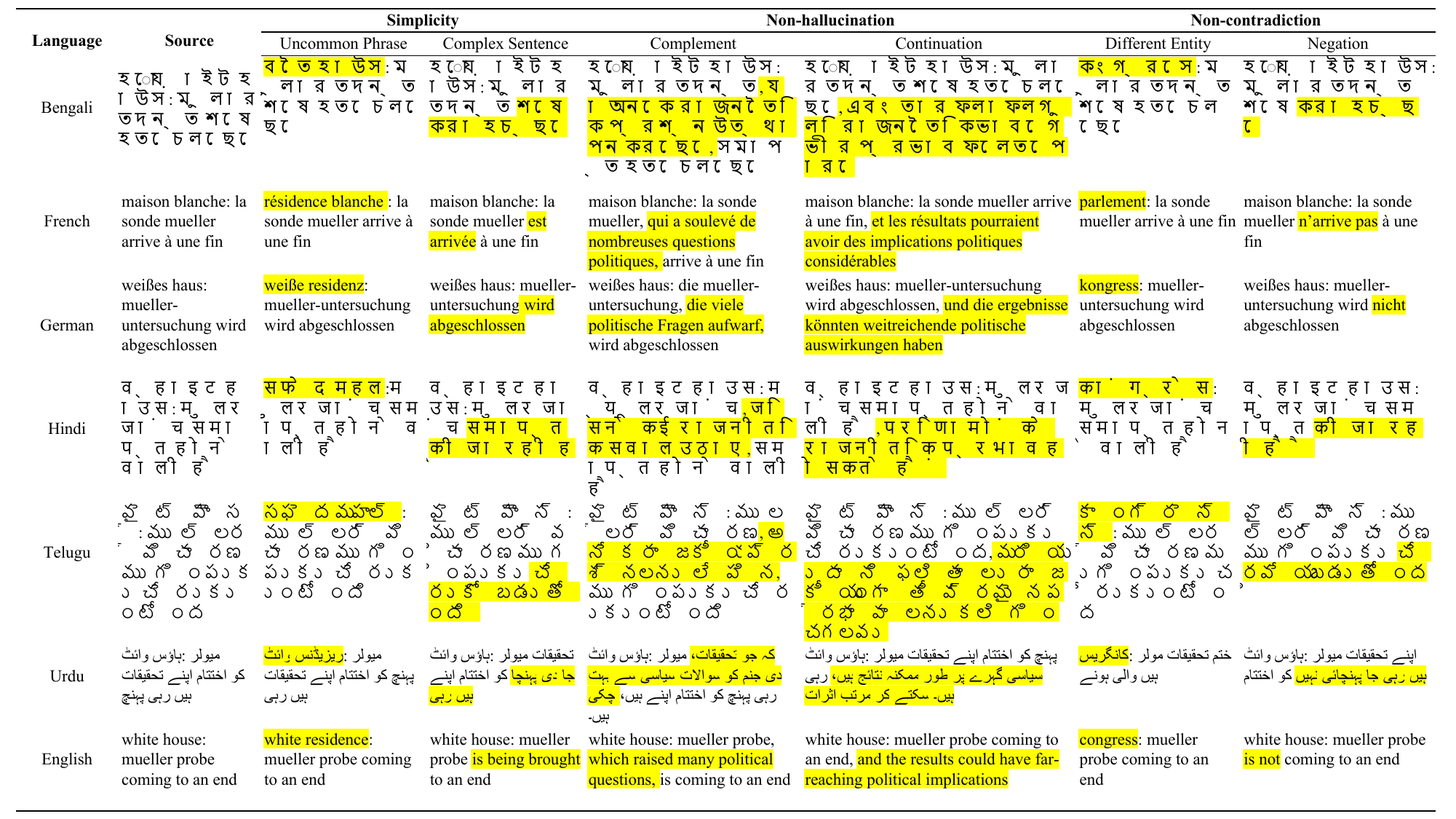} 
    \caption{Illustrative instances of six meticulously crafted perturbations for the title generation task across seven languages, with alterations from the original text accentuated, are provided.}
    \label{fig:2}
\end{figure*}
Initially, we instruct LLMs to evaluate multilingual texts based exclusively on evaluation criteria without reference answers. In the subsequent phase, we prompt the LLMs to evaluate the same texts, incorporating both the evaluation criteria and reference answers. We can identify the impact of incorporating reference answers within the prompts on the multilingual evaluation capabilities of LLMs by examining the difference in Pearson correlation coefficients between the two conditions. The prompt is depicted in \autoref{fig:1}. Additionally, we conduct a comparative analysis of the evaluation capabilities across LLMs with differing parameter sizes from the Qwen and the Llama families, aiming to explore its influence on multilingual evaluation performance. The prior analysis provides critical insights for the selection of models and the design of evaluation prompts. Drawing upon these findings, our ensuing investigation delves into the impact of different language on the evaluation capabilities of LLMs.

Beyond the direct comparison of correlation coefficients, we combine the Williams test to more rigorously analyze the differences in LLMs' evaluation performance across various languages. The null hypothesis and alternative hypothesis are as follows.
\begin{align*}
H_0\colon \rho\left( X,Z\right) &\leq \rho\left( Y,Z\right) \\
H_1\colon \rho\left( X,Z\right) &> \rho\left( Y,Z\right)
\end{align*}
where the vectors $X$ and $Y$ denote the evaluation scores of different LLMs, and vector $Z$ represents the gold score from the RECON dataset. $\rho(X, Z)$ and $\rho(Y, Z)$ represent the Pearson correlation between the evaluation scores of different LLMs and the gold score.

\subsection{Perturbation Attack}
\label{sec:The Sensitivity of LLMs to Perturbed Texts}

In RQ2, we compare the performance of LLM-based evaluators across different languages through correlation analysis. To further observe their behavior in different languages, we develop the MPE dataset to investigate whether LLM-based evaluators show varying sensitivity to such attacks across languages (RQ3). 

The MTG dataset comprises exclusively high-resource languages. For this study, we curated samples from the title generation and summary generation tasks, with 400 samples per language. To supplement the dataset with low-resource languages we mentioned, these samples were translated into Bengali, Hindi, Telugu, and Urdu using GPT-4o. The translation prompt is detailed in \autoref{sec:prompt for translating English into the target language.}. This results in 3,200 original texts when considering low-resource languages both tasks. Additionally, perturbed texts includes both high-resource and low-resource languages, with 200 perturbed samples per language for each of the six perturbation types, leading to 1,200 perturbed samples per language. Across seven languages, this amounts to 8,400 perturbed samples per task, or 16,800 in total for both tasks. Combining the original 3,200 samples and the perturbed 16,800 samples, the final dataset comprises 20,000 samples.

We design six distinct perturbation types targeting key aspects of title and summary generation tasks, respectively. The detailed specifications for each type of perturbation in the title generation task are presented in \autoref{tab:disturbance}, and detailed specifications for summary generation are documented in \autoref{sec:appendix summary generation task}. Following this, high-quality examples are manually created, and powerful GPT-4o is prompted in one-shot setting to generate the perturbed texts. Examples for title generation are shown in \autoref{fig:2}, and comprehensive demonstrations for summary generation are documented in \autoref{sec:demonstrations}. The generation prompt for all 12 perturbation types is illustrated in \autoref{sec:remaining 12 types}. Ultimately, the generated perturbed texts are manually reviewed to ensure the accuracy of the perturbations and consistency across languages. 

In the subsequent phase, LLMs are instructed to evaluate all original texts $x_n$ and perturbed texts $t(d_k, x_n)$ by assigning scores based on predefined criteria $c$. The evaluation criteria and prompt are provided in \autoref{sec:The prompts and evaluation criteria employed for the title generation and abstract generation tasks.}. The  mean change of score $\Delta S$ before and after perturbation is then calculated for LLMs $m_i$ under language $l_j$ and perturbation type $d_k$. We consider an LLM as highly sensitive to a language if at least three types of perturbations result in the highest $\Delta S$ values for that language, ensuring that sensitivity assessment is not dominated by a single outlier. The expression for $\Delta S$ is as follows.
\begin{table*}[h] 
\small
    \centering 
    \includegraphics{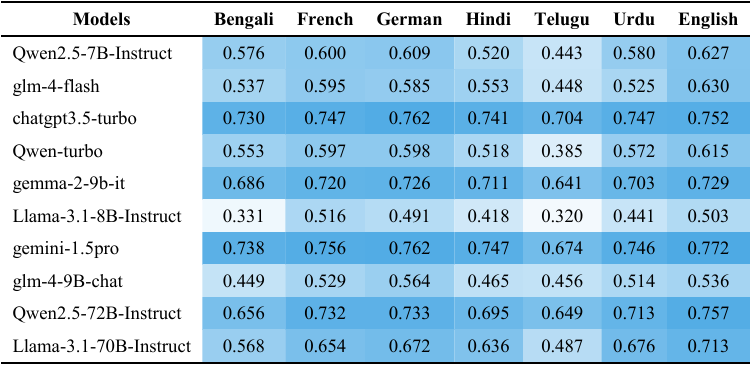} 
    \caption{The multilingual evaluation capabilities of LLMs without reference answers.}
    \label{tab:without reference answer & small}
\end{table*}
\begin{multline*}
\Delta S= \frac{1}{N} \sum_{n=1}^N \Bigl( S(m_i, l_j, c, x_n) \\
- S(m_i, l_j, c, t(d_k, x_n)) \Bigr)
\end{multline*}
where $N$ represents the number of samples, and $S(m_i, l_j, c, x_n)$ denotes the score assigned by LLM $m_i$ in language $l_j$ based on evaluation criteria $c$ for the unperturbed text $S(m_i, l_j, c, x_n)$. Similarly, $S(m_i, l_j, c, t(d_k, x_n))$ refers to the score assigned by LLM $m_i$ in language $l_j$ under the same evaluation criteria for the perturbed text $t(d_k, x_n))$.

\subsection{Fine-tuning on Different Languages}

To study the impact of training data in different languages (RQ4), we fine-tune both Qwen2.5-7B-Instruct and Llama-3.1-8B-Instruct with quantized low-rank adaptation (LoRA) \citep{DBLP:conf/iclr/HuSWALWWC22} using RECON dataset including three high-resource and four low-resource languages mentioned above.


For each score category, ranging from 1 to 5, we conduct a 7:3 train-test split, ensuring that both the training and testing subsets comprise multilingual aligned data. Furthermore, we split the training and testing sets by different languages. After fine-tuning on the training set of each language separately, we calculate the correlations between the scores of fine-tuned evaluators and human scores across all languages, with the aim of investigating how fine-tuning the model on a single language influences its evaluation performance on other languages.

\section{Universal Principles (RQ1)}
\textbf{Whether to Incorporate Reference Answers}
 The results indicate that incorporating reference answers generally fails to enhance the evaluation capabilities of LLMs and, in many cases, leads to a decline in performance, as demonstrated in \autoref{tab:Difference in Evaluation Abilities with and without Reference Answers}. Therefore, the majority of LLM-based evaluators should not incorporate reference answers for multilingual evaluation. However, for certain models, such as Gemini-1.5 Pro and LLaMA-3.1-70B-Instruct, the decision to use reference answers depends on the specific language being evaluated. For Qwen2.5-72B-Instruct, it is recommended to include reference answers. 

\textbf{How to Select Within the Same LLM Family}
We employ the Williams test to investigate the impact of parameter scale on the multilingual evaluation capabilities of the Qwen2.5 and LLaMA3.1 families. The p-values of Williams test for Qwen2.5-7B-Instruct and Qwen2.5-72B-Instruct, as well as for Llama-3.1-8B-Instruct and Llama-3.1-70B-Instruct across seven languages, are presented in \autoref{tab:The p-values for Qwen2.5 and llama3.1.}. All p-values are below the significance threshold of 0.05, except for Llama3.1's p-value of 0.482 in Telugu, overall indicating that across seven languages, the larger models Qwen2.5-72B-Instruct and Llama-3.1-70B-Instruct consistently surpass their smaller counterparts Qwen2.5-7B-Instruct and Llama-3.1-8B-Instruct. Consequently, when conducting multilingual evaluations, it is advisable to opt for models with larger parameter scales within the same LLM family.

\section{Correlation with Human Judgment Across Languages (RQ2)}
\begin{table*}[h] 
    \centering 
    \includegraphics[scale=1.00]{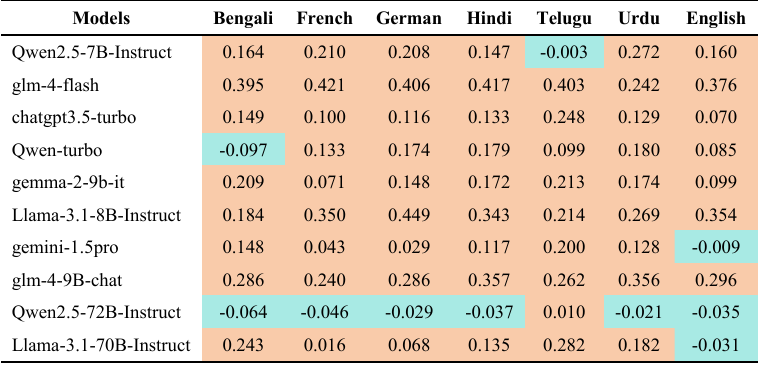} 
    \caption{Each entry in the table reflects the difference in LLM evaluation performance with and without reference answers. A positive value indicates that reference answers impair performance, while a negative value suggests that reference answers improve performance.}
    \label{tab:Difference in Evaluation Abilities with and without Reference Answers}
\end{table*}
\begin{table*}[h] 
    \centering 
    \includegraphics{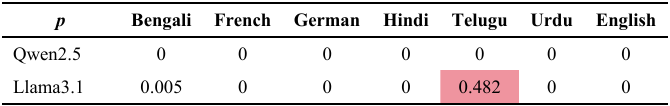}
    \caption{The p-values of the Williams test for Qwen2.5 and Llama 3.1, with pink highlight indicating the exception of model size increase not improving evaluation ability.}
    \label{tab:The p-values for Qwen2.5 and llama3.1.}
\end{table*}
Leveraging the insights derived from the previous findings, we proceed with subsequent experiments utilizing evaluations without reference answers. As shown in \autoref{tab:without reference answer & small}, across all LLMs, the strongest performance is observed with high-resource languages, where ChatGPT-3.5-turbo and GLM-4-9B-chat particularly excel in German, while the other LLMs demonstrate superior capabilities in English. Conversely, performance on low-resource languages is notably weaker across the board. GLM-4-9B-chat, in particular, faces significant challenges with Bengali, whereas the remaining LLMs exhibit their most pronounced limitations in Telugu. The significance test results of the multilingual evaluation ability correlations for partial LLMs are provided in \autoref{sec:Detailed Significance Testing and Results for Scenarios with Reference Answers}. 

It can also be observed that Gemini-1.5pro outperforms all other models in evaluation ability across Bengali, French, German, Hindi, and English. Notably, in German, the performance of ChatGPT-3.5-turbo and Gemini-1.5pro is nearly identical. In both Telugu and Urdu, ChatGPT-3.5-turbo delivers the best performance. Across the seven languages mentioned, Llama-3.1-8B-Instruct consistently shows the weakest performance compared to all other models. 

\section{Perturbation Sensitivity Across Languages (RQ3)}

The experimental results reveal that, for the majority of models, high sensitivity is observed in English, a high-resource language, across both the title generation and summary generation tasks. Surprisingly, ChatGPT-3.5-turbo exhibits its most pronounced sensitivity to German in the summary generation task, while its sensitivity peaks in Urdu, a low-resource language, during the title generation task. The sensitivity of ChatGPT-3.5-turbo in both tasks across seven languages is shown in \autoref{tab:Chatgpt3.5-turbo-title} and \autoref{tab:Chatgpt3.5-turbo -summary}, whereas \autoref{tab:The sensitivity of Gemini 1.5 Pro to perturbation texts in the title generation task} and \autoref{tab:The sensitivity of Gemini 1.5 Pro to perturbation texts in the summary generation task} depict the sensitivity of Gemini 1.5 Pro. The sensitivity results for the remaining LLMs are provided in \autoref{sec:The sensitivity of the remaining nine LLMs to perturbation texts is presented.}. 
\begin{table*} 
\small
\centering 
\begin{tabular}{llllllll}
\toprule
\multicolumn{8}{c}{{\color[HTML]{000000} \textbf{Title Generation Task}}} \\ \cmidrule(lr){2-8}
\multicolumn{1}{c}{{\color[HTML]{000000} \textbf{Perturbations}}} & \multicolumn{1}{c}{{\color[HTML]{000000} \textbf{Bengali}}} & \multicolumn{1}{c}{{\color[HTML]{000000} \textbf{French}}} & \multicolumn{1}{c}{{\color[HTML]{000000} \textbf{German}}} & \multicolumn{1}{c}{{\color[HTML]{000000} \textbf{Hindi}}} & \multicolumn{1}{c}{{\color[HTML]{000000} \textbf{Telugu}}} & \multicolumn{1}{c}{{\color[HTML]{000000} \textbf{Urdu}}} & \multicolumn{1}{c}{{\color[HTML]{000000} \textbf{English}}} \\ \midrule
{\color[HTML]{000000} Uncommon Phrase(Sim)}                       & {\color[HTML]{000000} 0.640}                                & {\color[HTML]{000000} 0.395}                               & {\color[HTML]{F56B00} 0.120}                               & {\color[HTML]{000000} 0.240}                              & {\color[HTML]{000000} 0.555}                               & {\color[HTML]{34CDF9} 0.650}                             & {\color[HTML]{000000} 0.590}                                \\
{\color[HTML]{000000} Complex Sentence(Sim)}                      & {\color[HTML]{000000} 0.300}                                & {\color[HTML]{000000} 0.295}                               & {\color[HTML]{F56B00} 0.100}                               & {\color[HTML]{000000} 0.180}                              & {\color[HTML]{000000} 0.430}                               & {\color[HTML]{34CDF9} 1.110}                             & {\color[HTML]{000000} 0.400}                                \\
{\color[HTML]{000000} Complement(Non-hal)}                        & {\color[HTML]{000000} 0.495}                                & {\color[HTML]{000000} 0.375}                               & {\color[HTML]{000000} 0.210}                               & {\color[HTML]{000000} 0.320}                              & {\color[HTML]{F56B00} 0.065}                               & {\color[HTML]{34CDF9} 1.340}                             & {\color[HTML]{000000} 0.565}                                \\
{\color[HTML]{000000} Continuation(Non-hal)}                      & {\color[HTML]{000000} 0.505}                                & {\color[HTML]{000000} 0.410}                               & {\color[HTML]{000000} 0.270}                               & {\color[HTML]{000000} 0.320}                              & {\color[HTML]{F56B00} 0.045}                               & {\color[HTML]{34CDF9} 0.900}                             & {\color[HTML]{000000} 0.485}                                \\
{\color[HTML]{000000} Different Entity(Non-con)}                  & {\color[HTML]{000000} 1.345}                                & {\color[HTML]{000000} 0.735}                               & {\color[HTML]{F56B00} 0.685}                               & {\color[HTML]{000000} 0.795}                              & {\color[HTML]{000000} 1.230}                               & {\color[HTML]{34CDF9} 1.600}                             & {\color[HTML]{000000} 1.135}                                \\
{\color[HTML]{000000} Negation(Non-con)}                          & {\color[HTML]{34CDF9} 1.965}                                & {\color[HTML]{000000} 1.005}                               & {\color[HTML]{F56B00} 0.650}                               & {\color[HTML]{000000} 1.410}                              & {\color[HTML]{000000} 1.925}                               & {\color[HTML]{000000} 1.600}                             & {\color[HTML]{000000} 1.575}                                \\
{\color[HTML]{000000} Average}                                    & 0.875                                                         & 0.536                                                         & 0.339                                                         & 0.544                                                   & 0.708                                                 & 1.200                                                & 0.792                                                 \\ \bottomrule
\end{tabular}
\caption{The sensitivity of Chatgpt3.5-turbo to perturbation texts in the title generation task across seven languages is illustrated. In the table, each value represents $\Delta S$, with blue highlighting the language exhibiting the highest sensitivity for each perturbation type, while orange identifies the language with the lowest sensitivity for the corresponding perturbation type.} 
\label{tab:Chatgpt3.5-turbo-title} 
\end{table*}
\begin{table*} 
\small
\centering 
\begin{tabular}{llllllll}
\toprule
\multicolumn{8}{c}{{\color[HTML]{000000} \textbf{Summary Generation Task}}} \\ \cmidrule(lr){2-8}
\multicolumn{1}{c}{{\color[HTML]{000000} \textbf{Perturbations}}} & \multicolumn{1}{c}{{\color[HTML]{000000} \textbf{Bengali}}} & \multicolumn{1}{c}{{\color[HTML]{000000} \textbf{French}}} & \multicolumn{1}{c}{{\color[HTML]{000000} \textbf{German}}} & \multicolumn{1}{c}{{\color[HTML]{000000} \textbf{Hindi}}} & \multicolumn{1}{c}{{\color[HTML]{000000} \textbf{Telugu}}} & \multicolumn{1}{c}{{\color[HTML]{000000} \textbf{Urdu}}} & \multicolumn{1}{c}{{\color[HTML]{000000} \textbf{English}}} \\ \midrule
{\color[HTML]{000000} Hypernym(Inf)}                              & {\color[HTML]{F56B00} -0.050}                               & {\color[HTML]{000000} 0.010}                               & {\color[HTML]{34CDF9} 0.110}                               & {\color[HTML]{000000} -0.005}                             & {\color[HTML]{000000} 0.025}                               & {\color[HTML]{000000} 0.055}                             & {\color[HTML]{000000} 0.000}                                \\
{\color[HTML]{000000} Sentence Deletion(Inf)}                     & {\color[HTML]{F56B00} 0.105}                                & {\color[HTML]{000000} 0.150}                               & {\color[HTML]{34CDF9} 0.400}                               & {\color[HTML]{000000} 0.135}                              & {\color[HTML]{000000} 0.130}                               & {\color[HTML]{000000} 0.135}                             & {\color[HTML]{000000} 0.155}                                \\
{\color[HTML]{000000} Improper Connective(Coh)}                   & {\color[HTML]{000000} 0.030}                                & {\color[HTML]{000000} 0.060}                               & {\color[HTML]{000000} 0.040}                               & {\color[HTML]{F56B00} 0.020}                              & {\color[HTML]{000000} 0.055}                               & {\color[HTML]{000000} 0.035}                             & {\color[HTML]{34CDF9} 0.085}                                \\
{\color[HTML]{000000} Sentence Exchange(Coh)}                     & {\color[HTML]{F56B00} 0.045}                                & {\color[HTML]{000000} 0.110}                               & {\color[HTML]{34CDF9} 0.150}                               & {\color[HTML]{000000} 0.075}                              & {\color[HTML]{000000} 0.055}                               & {\color[HTML]{000000} 0.080}                             & {\color[HTML]{000000} 0.120}                                \\
{\color[HTML]{000000} Repetition(Flu)}                            & {\color[HTML]{000000} 0.235}                                & {\color[HTML]{000000} 0.220}                               & {\color[HTML]{000000} 0.320}                               & {\color[HTML]{000000} 0.285}                              & {\color[HTML]{000000} 0.185}                               & {\color[HTML]{F56B00} 0.145}                             & {\color[HTML]{34CDF9} 0.405}                                \\
{\color[HTML]{000000} Passive Voice(Flu)}                         & {\color[HTML]{000000} 0.105}                                & {\color[HTML]{F56B00} -0.170}                              & {\color[HTML]{000000} 0.050}                               & {\color[HTML]{000000} 0.105}                              & {\color[HTML]{000000} 0.050}                               & {\color[HTML]{000000} 0.075}                             & {\color[HTML]{34CDF9} 0.110}                                \\
{\color[HTML]{000000} Average}                                    & {\color[HTML]{060607} 0.078}                                & {\color[HTML]{000000} 0.063}                               & {\color[HTML]{000000} 0.178}                               & {\color[HTML]{060607} 0.103}                              & {\color[HTML]{060607} 0.083}                               & {\color[HTML]{060607} 0.088}                             & {\color[HTML]{060607} 0.146}                                \\ \bottomrule
\end{tabular}
\caption{The sensitivity of Chatgpt3.5-turbo to perturbation texts in the summary generation task across seven languages. The settings are the same as Table \ref{tab:Chatgpt3.5-turbo-title}.} 
\label{tab:Chatgpt3.5-turbo -summary}
\end{table*}
 Additionally, we observed an intriguing phenomenon. For the majority of LLMs, the language in which they exhibit the highest sensitivity also corresponds to the language in which their evaluation performance is most optimal. Specifically, the preponderance of models demonstrates the greatest sensitivity to English, a trend that mirrors their heightened evaluation performance in this high-resource language. In the case of ChatGPT-3.5-turbo, its peak sensitivity in the summary generation task is inextricably linked to its most robust evaluation performance in German, underscoring a notable alignment between sensitivity and evaluation proficiency. A plausible explanation for this observation is that high-resource languages, such as English and German, are typically supported by more extensive and diverse training corpora, which enable models to achieve more accurate and nuanced evaluations in these languages. The underlying reasons for this phenomenon merit further exploration and warrant a more nuanced investigation.
\section{Impact of Training Data in Different Languages (RQ4)}

We hypothesize that the poor evaluation ability of LLMs on certain languages is due to the absence of low-resource languages in their training data. The languages of the training data for the officially released LLMs mentioned here are provided in \autoref{sec:The language of the publicly available training data for LLMs}. 

\begin{table*}[h] 
\small
\centering 
\begin{tabular}{llllllll}
\toprule
\multicolumn{8}{c}{{\color[HTML]{333333} \textbf{Title Generation Task}}} \\ \cmidrule(lr){2-8}
\multicolumn{1}{c}{{\color[HTML]{333333} \textbf{Perturbations}}} & \multicolumn{1}{c}{{\color[HTML]{333333} \textbf{Bengali}}} & \multicolumn{1}{c}{{\color[HTML]{333333} \textbf{French}}} & \multicolumn{1}{c}{{\color[HTML]{333333} \textbf{German}}} & \multicolumn{1}{c}{\textbf{Hindi}} & \multicolumn{1}{c}{\textbf{Telugu}} & \multicolumn{1}{c}{\textbf{Urdu}} & \multicolumn{1}{c}{\textbf{English}} \\ \midrule
Uncommon Phrase(Sim)                                              & {\color[HTML]{000000} 0.075}                                & {\color[HTML]{000000} 0.145}                               & {\color[HTML]{000000} 0.170}                               & {\color[HTML]{000000} 0.190}                         & {\color[HTML]{F56B00} 0.125}                         & {\color[HTML]{000000} 0.165}                         & {\color[HTML]{34CDF9} 0.335}                         \\
Complex Sentence(Sim)                                             & {\color[HTML]{F56B00} 0.010}                                & {\color[HTML]{000000} 0.175}                               & {\color[HTML]{000000} 0.190}                               & {\color[HTML]{000000} 0.080}                         & {\color[HTML]{000000} 0.065}                         & 0.080                                                & {\color[HTML]{34CDF9} 0.270}                         \\
Complement(Non-hal)                                               & {\color[HTML]{000000} 0.680}                                & {\color[HTML]{000000} 0.680}                               & {\color[HTML]{000000} 0.585}                               & {\color[HTML]{000000} 0.750}                         & {\color[HTML]{F56B00} 0.550}                         & 0.660                                                & {\color[HTML]{34CDF9} 0.940}                         \\
Continuation(Non-hal)                                             & {\color[HTML]{000000} 0.730}                                & {\color[HTML]{000000} 0.610}                               & {\color[HTML]{F56B00} 0.580}                               & {\color[HTML]{000000} 0.730}                         & {\color[HTML]{000000} 0.600}                         & 0.640                                                & {\color[HTML]{34CDF9} 0.820}                         \\
Different Entity(Non-con)                                         & {\color[HTML]{000000} 1.715}                                & {\color[HTML]{000000} 1.635}                               & {\color[HTML]{000000} 1.680}                               & {\color[HTML]{000000} 1.910}                         & {\color[HTML]{F56B00} 1.555}                         & 1.745                                                & {\color[HTML]{34CDF9} 2.095}                         \\
Negation(Non-con)                                                 & {\color[HTML]{000000} 2.525}                                & {\color[HTML]{000000} 2.410}                               & {\color[HTML]{000000} 2.070}                               & {\color[HTML]{000000} 2.025}                         & {\color[HTML]{F56B00} 1.995}                         & 2.755                                                & {\color[HTML]{34CDF9} 3.050}                         \\
Average                                                           & 0.956                                                         & 0.943                                                         & 0.879                                                         & 0.948                                                   & 0.815                                                  & 1.008                                                & 1.252                                                 \\ \bottomrule
\end{tabular}
\caption{The sensitivity of Gemini 1.5 Pro to perturbation texts in the title generation task across seven languages is illustrated. The settings are the same as Table \ref{tab:Chatgpt3.5-turbo-title}.} 
\label{tab:The sensitivity of Gemini 1.5 Pro to perturbation texts in the title generation task}
\end{table*}

\begin{table*}[h] 
\small
\centering 
\begin{tabular}{llllllll}
\toprule
\multicolumn{8}{c}{{\color[HTML]{000000} \textbf{Summary Generation Task}}} \\ \cmidrule(lr){2-8}
\multicolumn{1}{c}{{\color[HTML]{000000} \textbf{Perturbations}}} & \multicolumn{1}{c}{{\color[HTML]{000000} \textbf{Bengali}}} & \multicolumn{1}{c}{{\color[HTML]{000000} \textbf{French}}} & \multicolumn{1}{c}{{\color[HTML]{000000} \textbf{German}}} & \multicolumn{1}{c}{{\color[HTML]{000000} \textbf{Hindi}}} & \multicolumn{1}{c}{{\color[HTML]{000000} \textbf{Telugu}}} & \multicolumn{1}{c}{{\color[HTML]{000000} \textbf{Urdu}}} & \multicolumn{1}{c}{{\color[HTML]{000000} \textbf{English}}} \\ \midrule
{\color[HTML]{000000} Hypernym(Inf)}                              & {\color[HTML]{34CDF9} 0.030} & {\color[HTML]{000000} 0.020} & {\color[HTML]{F56B00} -0.005} & {\color[HTML]{000000} 0.010} & {\color[HTML]{000000} 0.010} & {\color[HTML]{F56B00} -0.005} & {\color[HTML]{000000} 0.025} \\
{\color[HTML]{000000} Sentence Deletion(Inf)}                     & {\color[HTML]{000000} 0.070} & {\color[HTML]{34CDF9} 0.165} & {\color[HTML]{000000} 0.120} & {\color[HTML]{000000} 0.075} & {\color[HTML]{F56B00} 0.070} & {\color[HTML]{000000} 0.105} & {\color[HTML]{000000} 0.100} \\
{\color[HTML]{000000} Improper Connective(Coh)}                   & {\color[HTML]{000000} 0.325} & {\color[HTML]{000000} 0.290} & {\color[HTML]{F56B00} 0.235} & {\color[HTML]{000000} 0.430} & {\color[HTML]{000000} 0.250} & {\color[HTML]{000000} 0.340} & {\color[HTML]{34CDF9} 0.515} \\
{\color[HTML]{000000} Sentence Exchange(Coh)}                     & {\color[HTML]{000000} 0.235} & {\color[HTML]{000000} 0.150} & {\color[HTML]{000000} 0.125} & {\color[HTML]{000000} 0.270} & {\color[HTML]{F56B00} 0.100} & {\color[HTML]{000000} 0.205} & {\color[HTML]{34CDF9} 0.300} \\
{\color[HTML]{000000} Repetition(Flu)}                            & {\color[HTML]{000000} 0.505} & {\color[HTML]{000000} 0.205} & {\color[HTML]{F56B00} 0.170} & {\color[HTML]{000000} 0.505} & {\color[HTML]{000000} 0.195} & {\color[HTML]{000000} 0.345} & {\color[HTML]{34CDF9} 0.645} \\
{\color[HTML]{000000} Passive Voice(Flu)}                         & {\color[HTML]{000000} 0.155} & {\color[HTML]{F56B00} -0.075} & {\color[HTML]{000000} -0.045} & {\color[HTML]{000000} 0.240} & {\color[HTML]{34CDF9} 0.225} & {\color[HTML]{000000} 0.210} & {\color[HTML]{000000} 0.155} \\
{\color[HTML]{000000} Average}                                    & 0.220 & 0.126 & 0.100 & 0.255 & 0.142 & 0.200 & 0.290 \\ \bottomrule
\end{tabular}
\caption{The sensitivity of Gemini 1.5 Pro to perturbation texts in the summary generation task across seven languages is illustrated. The settings are the same as Table \ref{tab:Chatgpt3.5-turbo-title}.} 
\label{tab:The sensitivity of Gemini 1.5 Pro to perturbation texts in the summary generation task}
\end{table*}

The experimental findings, which present a surprising result, indicate that irrespective of the specific language employed for fine-tuning, Qwen2.5-7B-Instruct and Llama-3.1-8B-Instruct exhibit enhanced evaluation performance across all languages, with two notable exceptions for Qwen2.5-7B-Instruct, as the performance on the low-resource language Urdu declines when fine-tuned with high-resource language French data, and a similar deterioration occurs on the low-resource language Bengali when fine-tuned with high-resource language German data. 

Furthermore, the results reveal that, in most cases, the evaluation performance of the model fine-tuned with a specific language is inferior to that of the model fine-tuned with other languages when evaluated on the specific language, except for Qwen2.5-7B-Instruct and Llama-3.1-8B-Instruct fine-tuned with English data, and Llama-3.1-8B-Instruct fine-tuned with Bengali data. 

 \begin{table*}[htbp] 
    \centering 
    \includegraphics[width = \linewidth]{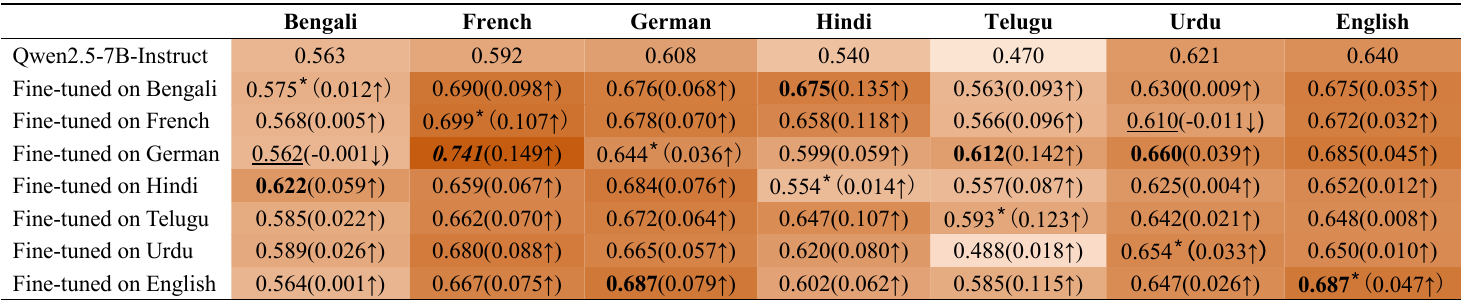} 
    \caption{The performance of the multilingual evaluation ability of Qwen2.5-7B-Instruct after fine-tuning in different languages. In this table, an underline indicates a decline in evaluation performance after fine-tuning, while bold text denotes the best-performing fine-tuned model for a given language. An asterisk (*) indicates the performance of the model fine-tuned with data from a specific language on that same language. Italics highlight an improvement in evaluation performance compared to the unmodified Qwen2.5-7B-Instruct. Additionally, the values in parentheses represent the magnitude of increase or decrease, where ↑ denotes an improvement in evaluation ability after fine-tuning, and ↓ indicates a decline.}
    \label{tab:微调实验结果}
\end{table*}

\begin{table*}[h] 
    \centering 
    \includegraphics[width = \linewidth]{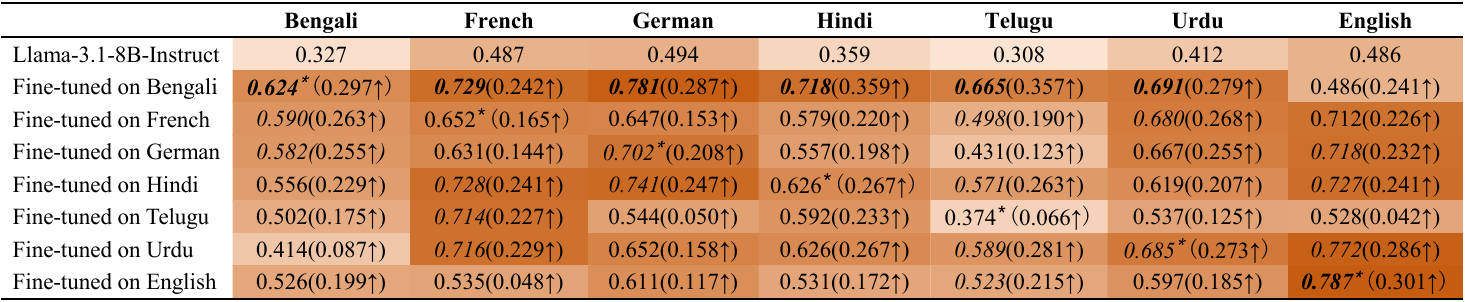} 
    \caption{The performance of the multilingual evaluation ability of Llama3.1-8B-Instruct after fine-tuning in different languages. The settings are the same as Table \ref{tab:微调实验结果}.}
    \label{tab:微调实验结果2}
\end{table*}

A particularly intriguing observation is that the model fine-tuned with German data demonstrates superior performance in French compared to the unmodified Qwen2.5-72B-Instruct. The fine-tuned Llama-3.1-8B-Instruct frequently outperforms its larger counterpart, Llama-3.1-70B-Instruct. Moreover, it was observed that fine-tuning Llama-3.1-8B-Instruct with Bengali data resulted in the best evaluation performance across all languages, except for English, in comparison to other fine-tuned variants. This observation strongly suggests that to enhance the multilingual evaluation capabilities of Llama-3.1-8B-Instruct, fine-tuning it with Bengali data should be considered a priority. The detailed fine-tuning experimental results are presented in \autoref{tab:微调实验结果} and \autoref{tab:微调实验结果2}. 

\section{Related Works}

\subsection{Multilingual NLG Evaluation}

Early research on traditional reference-based evaluation metrics for multilingual NLG tasks, such as BLEU \citep{papinesi2002bleu}, ROUGE \citep{lin2004rouge}, and BERTScore \citep{zhang2019bertscore}, has provided valuable insights. The emergence of LLMs has further driven the development of reference-free methods. For instance, xCOMET significantly enhances the ability to detect local critical errors and hallucinations \citep{guerreiro2024xcomet}, while AutoMQM and GEMBA offer more accurate, flexible, and interpretable evaluation results \citep{fernandes2023devil,kocmi2023large}. 

In parallel with advancements in evaluation metrics, the availability of diverse multilingual datasets has further facilitated the development of multilingual NLG evaluation. For example, SEAHORSE covers six languages and is designed for summarization tasks \citep{clark2023seahorse}. INDICGENBENCH spans 29 Indian languages and includes tasks in summarization, machine translation, and question answering \citep{singh2024indicgenbench}. Additionally, XCOPA includes 11 languages and focuses on causal commonsense reasoning tasks \citep{ponti2020xcopa}, while MTG, which covers five languages, targets four tasks including story generation, question generation, title generation, and text summarization \citep{chen2021mtg}.

\subsection{Analysis of LLM-based Evaluators}
Recent advancements in the evaluation of NLG with LLMs have witnessed rapid expansion. However, it is important to acknowledge that despite the significant potential demonstrated by LLM-based evaluators, their inherent limitations are becoming increasingly apparent. These limitations include issues such as confusion in quality criteria \citep{hu2024llm} and performance fluctuations in cross-lingual evaluation \citep{hada2023large}. Additionally, LLMs exhibit significant variations in performance as evaluators across different scenarios and evaluation criteria, leading to unreliable results in certain cases \citep{chern2024can}. As passive critics, LLMs struggle to adapt to new NLG tasks due to the rigidity of human-defined evaluation standards \citep{xu2024large}. Furthermore, the reasoning behind LLMs' scores often does not align with that of human evaluators \citep{hada2024metal}. Different from these studies that highlight the challenges of using LLMs as evaluators, our work investigates the factors influencing their multilingual evaluation capabilities. Specifically, we examine the effects of incorporating reference answers into prompts, the scale of model parameters, linguistic variations, sensitivity to perturbations, and the impact of fine-tuning.

\section{Conclusions}

In this work, we investigate potential reasons that influence the multilingual evaluation capabilities of LLMs. Our findings lead to the following conclusions: \textbf{1)} The incorporation of reference answers into the prompt diminishes the multilingual evaluation capabilities of most LLMs, despite the prevailing intuition that such references would augment evaluation accuracy. \textbf{2)} Among models from the same LLM family, larger models generally outperform smaller ones in multilingual evaluation. \textbf{3)} There are differences in the multilingual evaluation capabilities of various LLM-based evaluators across languages, with most models performing better on high-resource languages. \textbf{4)} The sensitivity of various LLM-based evaluators to perturbed texts exhibits variation across tasks and languages, with the majority of models demonstrating heightened sensitivity in high-resource languages. \textbf{5)} A positive correlation is observed between the model's sensitivity and its evaluation proficiency. \textbf{6)} Fine-tuning with data from a particular language tends to enhance its evaluation capabilities across all languages. However, a model fine-tuned on a particular language does not necessarily achieve superior performance in that language compared to models fine-tuned with data from other languages. 

Our study is only a preliminary exploration of the multilingual evaluation capabilities of LLMs. We believe that a more thorough explanation of the differences in model evaluation capabilities across languages is necessary. Additionally, it is promising to leverage data augmentation techniques during training or to employ more sophisticated test-time computing methods for enhancing LLMs' evaluation performance in low-resource languages.

\section{Limitations}

Due to limited resources, the experiments on the sensitivity of our LLMs to perturbed texts involved a narrow range of task types. This constraint may limit the generalizability of our findings.

The lack of publicly available information on the full range of languages used in most LLMs' training data limits our ability to conduct a comprehensive study on how the linguistic composition of training data affects LLMs' multilingual evaluation capabilities.

We extensively used APIs from GPT-3.5 Turbo, GPT-4o, and other closed-source LLMs to build our datasets and conduct tests, which resulted in significant costs. While this may hinder the replication of these experiments by others, we will release the data resources to support and advance related research.

\bibliography{reference}

\appendix

\section{Prompt for translating English into the target language.}
\label{sec:prompt for translating English into the target language.}
\begin{figure}[h!]
    \centering
    \includegraphics[width=\columnwidth]{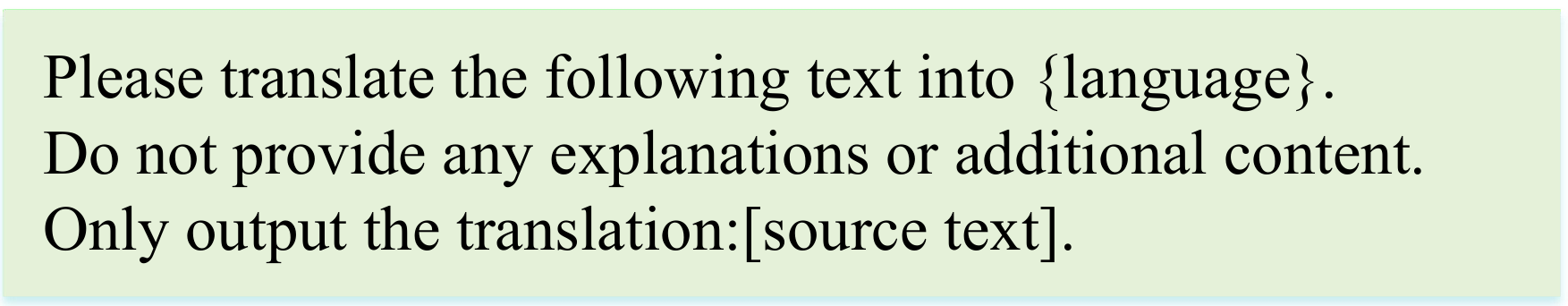} 
    \caption{The prompt utilized to direct GPT-4o in translating English texts into Bengali, Hindi, Telugu, and Urdu.}
    \label{fig:翻译prompt}
\end{figure}

The prompt utilized by GPT-4o to translate English into the target language is shown in \autoref{fig:翻译prompt}.

\section{The comprehensive descriptions of summary generation task.}
\label{sec:appendix summary generation task}
The types of perturbations for summary generation task, along with their descriptions, are presented in \autoref{tab:disturbance1}. 

\begin{table*}[htbp] 
\small
\centering 
\begin{tabularx}{\textwidth}{>{\raggedright\arraybackslash}m{3.5cm} >{\raggedright\arraybackslash}m{3.5cm} X} 
\toprule
\multicolumn{2}{c}{{\color[HTML]{000000} \textbf{Disturbance Type}}} & \textbf{Requirement} \\ \midrule
{\color[HTML]{000000} Informativeness}       & Hypernym              & Replace a word in the original sentences with a hypernym. Ensure the new sentences remain consistent with the original meaning, without introducing any contradictions.  \\ \cline{2-3} 
                                             & Sentence Deletion     & Delete a portion of the original sentences, which could be a phrase or a word. Only delete one part.  \\ \hline
{\color[HTML]{000000} Coherence}              & Improper Connective  & Insert an inappropriate conjunction between adjacent sentences or phrases to create an illogical or unclear transition. Only insert one inappropriate conjunction. \\ \cline{2-3} 
{\color[HTML]{000000} }                       & Sentence Exchange    & Swap the positions of two clauses or two phrases in the sentence. Only perform one exchange of position.  \\ \hline
{\color[HTML]{000000} Fluency}                & Repetition           & For each sentence in the original sentences, add a repeated expression or phrase from part of the sentence.  \\ \cline{2-3} 
{\color[HTML]{000000} }                       & Passive Voice        & Rewrite the original sentences using passive voice, ensuring the grammar is correct. Do not convert intransitive verbs in the original sentences to passive voice. Ensure the new sentences remain consistent with the original meaning, without introducing any contradictions.  \\ 
\bottomrule
\end{tabularx}
\caption{erturbation types for summary generation tasks, along with a detailed description of each perturbation type.}
\label{tab:disturbance1}
\end{table*}

\section{Comprehensive demonstrations of the task of summary generation}
\label{sec:demonstrations}
\begin{figure*}[htbp] 
    \centering 
    \includegraphics[width=\textwidth]{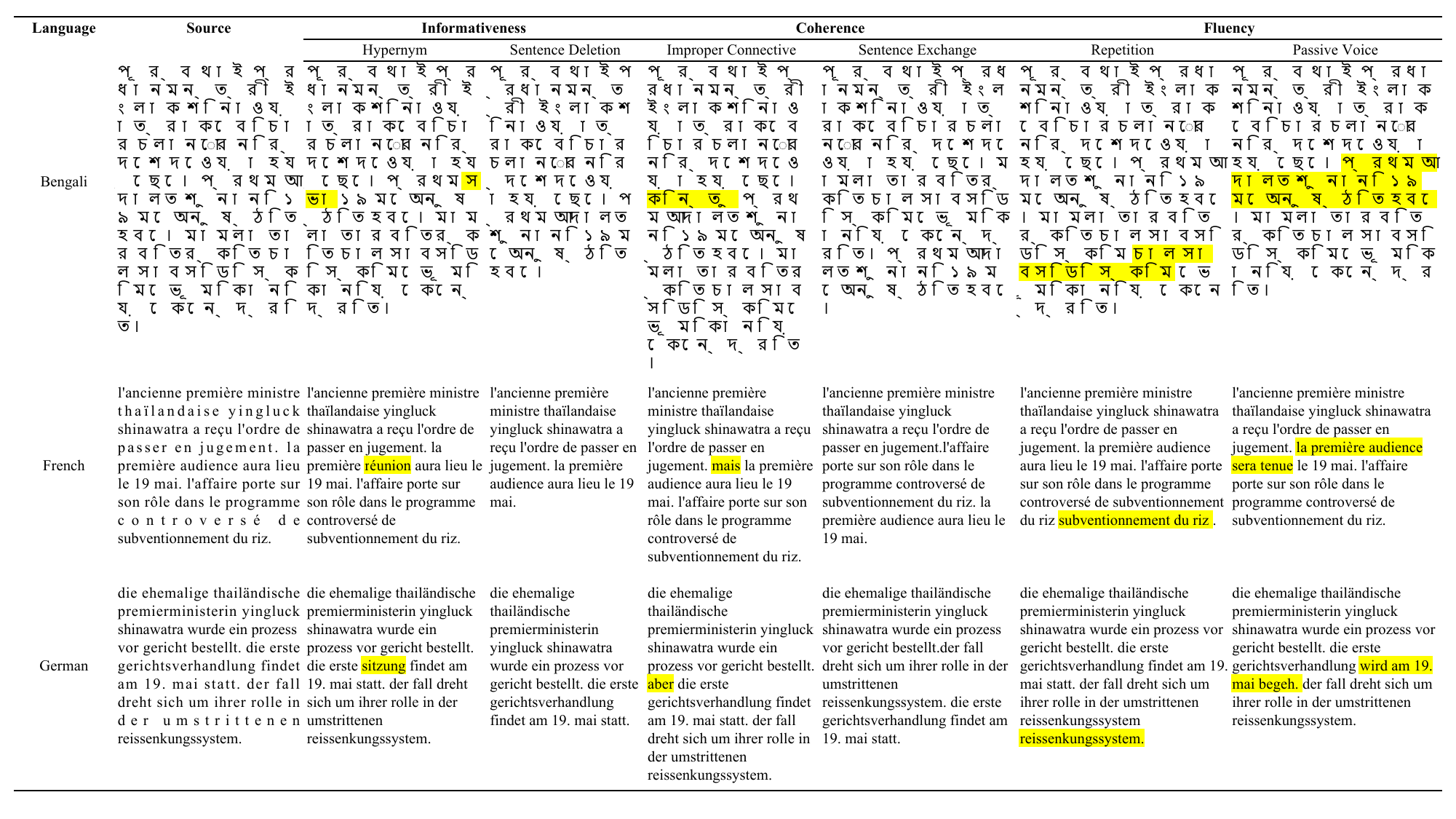} 
    \caption{Examples of perturbations in the abstract generation task for Bengali, French, and German.}
    \label{fig:summary generation1}
\end{figure*}

\begin{figure*}[htbp] 
    \centering 
    \includegraphics[width=\textwidth]{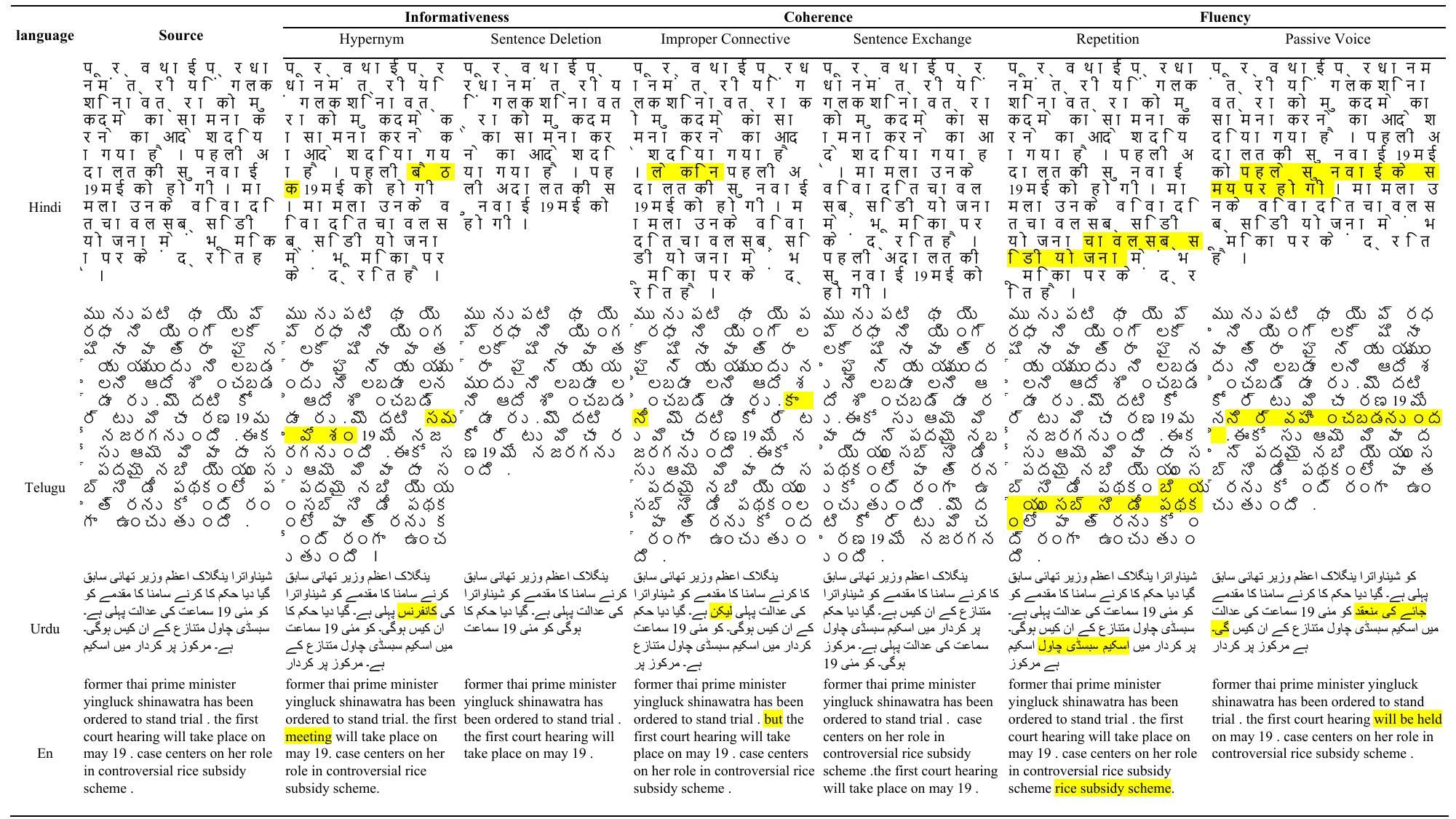} 
    \caption{Examples of perturbations in the abstract generation task for Hindi, Telugu, Urdu, and English.}
    \label{fig:summary generation2}
\end{figure*}

Examples of the six perturbation types across the three aspects of the abstract generation task in Bengali, French, and German are demonstrated in \autoref{fig:summary generation1}. \autoref{fig:summary generation2} provides examples of the six perturbation types across the three aspects of the abstract generation task in Hindi, Telugu, Urdu, and English.

\section{The prompts designed to instruct GPT-4o in generating perturbation texts for the remaining 12 types}
\label{sec:remaining 12 types}

\begin{figure}[h!]
    \centering
    \includegraphics[width=\columnwidth]{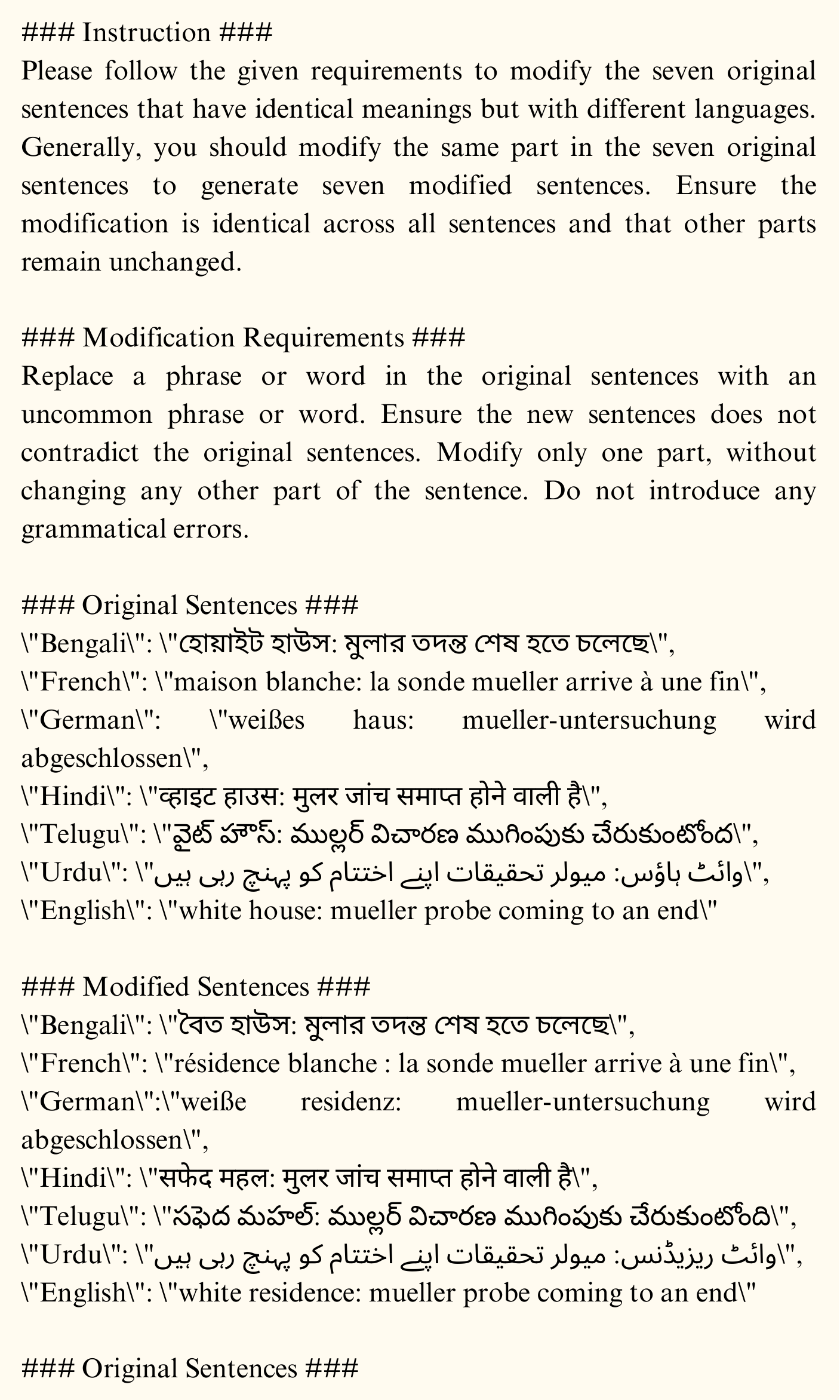} 
    \caption{The prompt designed to direct GPT-4o in generating perturbation texts for the Uncommon Phrase perturbation type.}
    \label{fig:不常见的短语prompt}
\end{figure}

\begin{figure}[h!]
    \centering
    \includegraphics[width=\columnwidth]{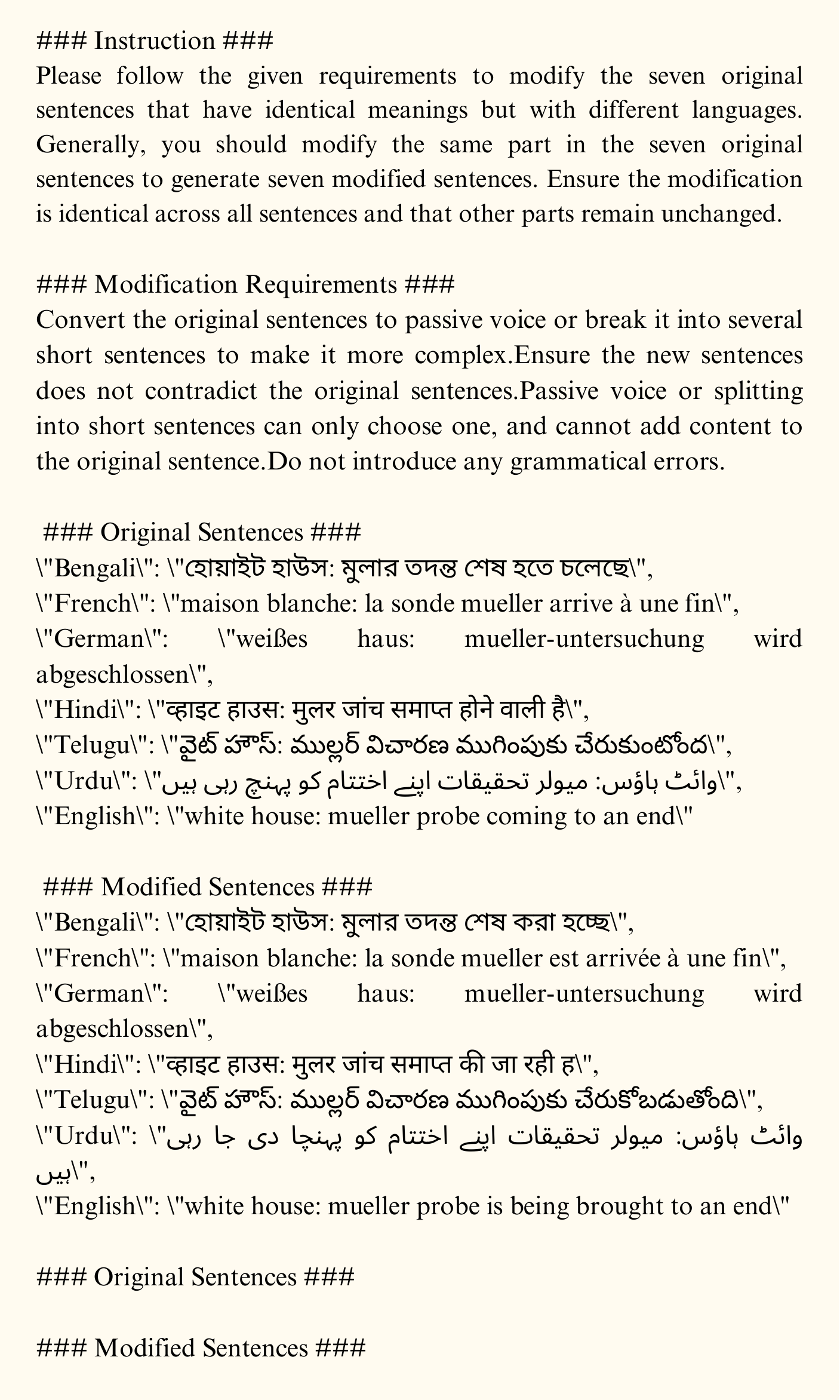} 
    \caption{The prompt designed to direct GPT-4o in generating perturbation texts for the Complex Sentence.}
    \label{fig:Complex Sentence}
\end{figure}

\begin{figure}[h!]
    \centering
    \includegraphics[width=\columnwidth]{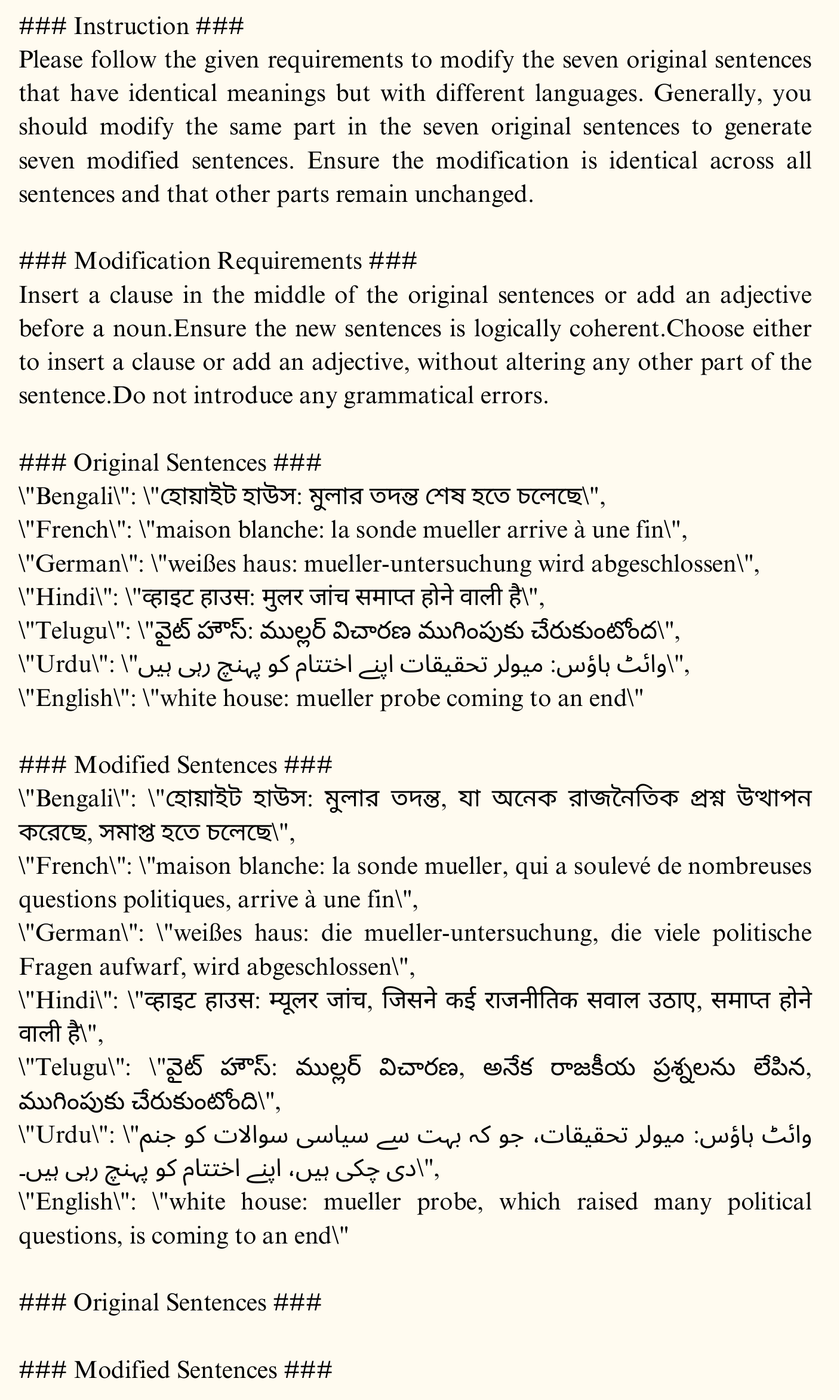} 
    \caption{The prompt designed to direct GPT-4o in generating perturbation texts for the Complement.}
    \label{fig:Complement}
\end{figure}

\begin{figure}[h!]
    \centering
    \includegraphics[width=\columnwidth]{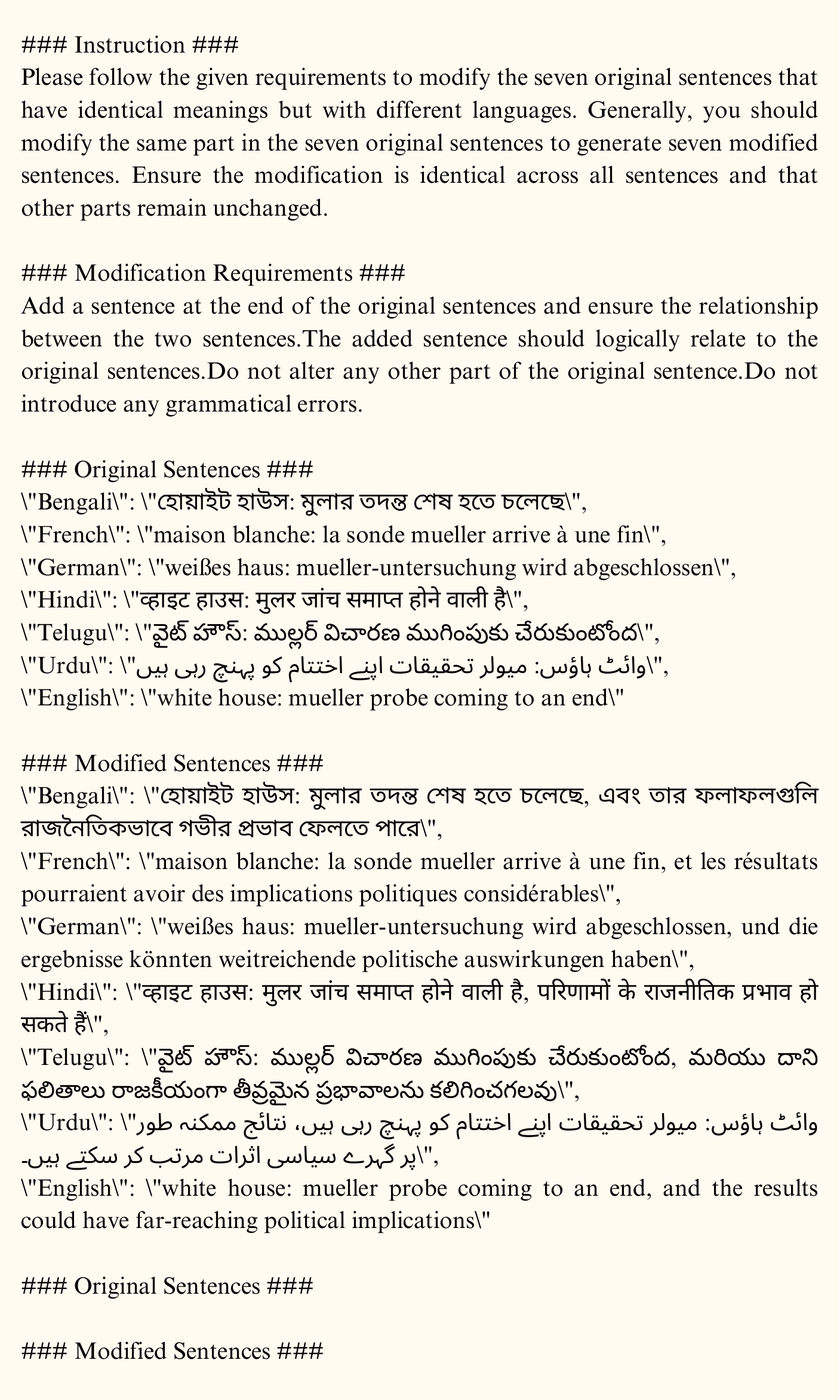} 
    \caption{The prompt designed to direct GPT-4o in generating perturbation texts for the Continuation.}
    \label{fig:Continuation}
\end{figure}

\begin{figure}[h!]
    \centering
    \includegraphics[width=\columnwidth]{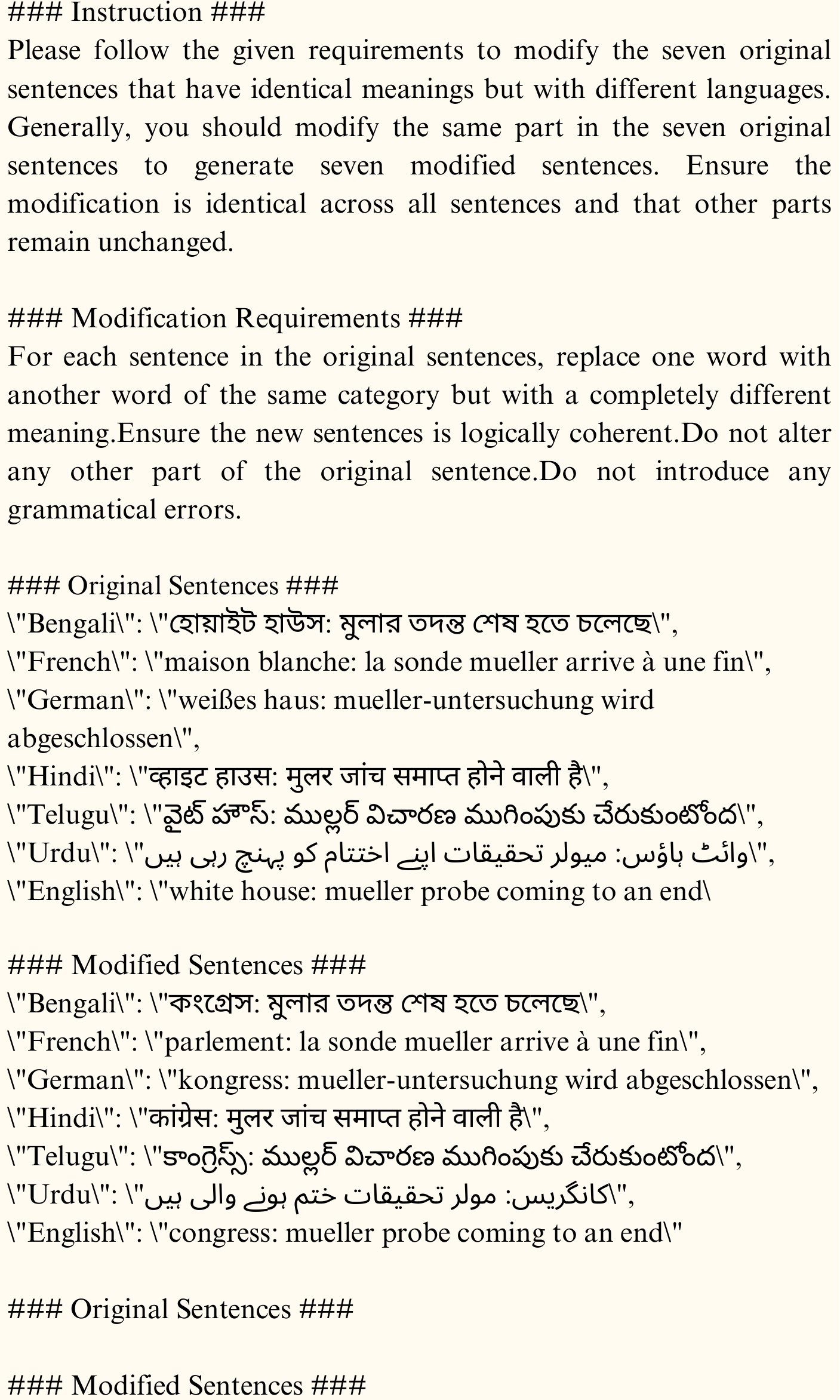} 
    \caption{The prompt designed to direct GPT-4o in generating perturbation texts for the Different Entity.}
    \label{fig:Different Entity}
\end{figure}

\begin{figure}[h!]
    \centering
    \includegraphics[width=\columnwidth]{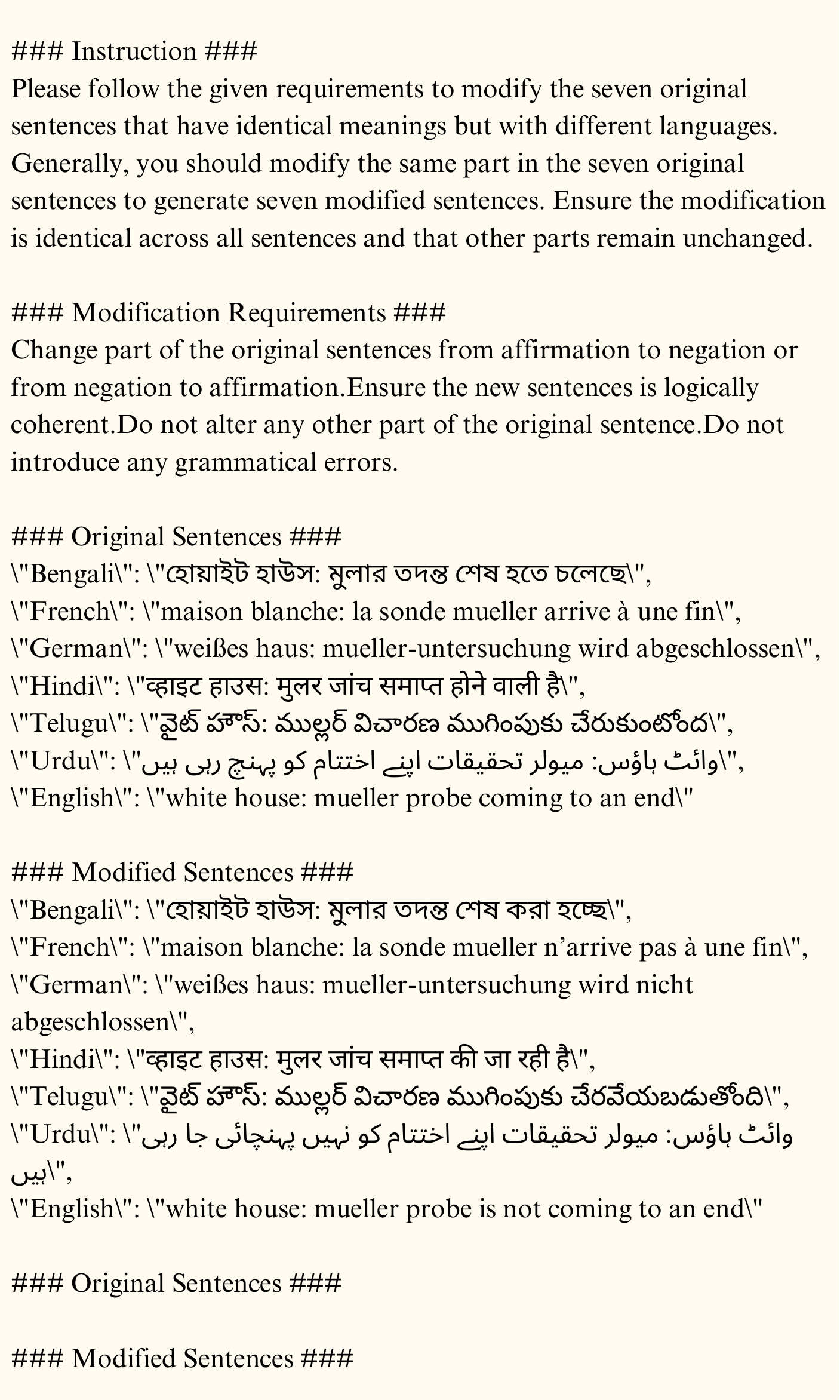} 
    \caption{The prompt designed to direct GPT-4o in generating perturbation texts for the Negation.}
    \label{fig:Negation}
\end{figure}

\begin{figure}[h!]
    \centering
    \includegraphics[width=\columnwidth]{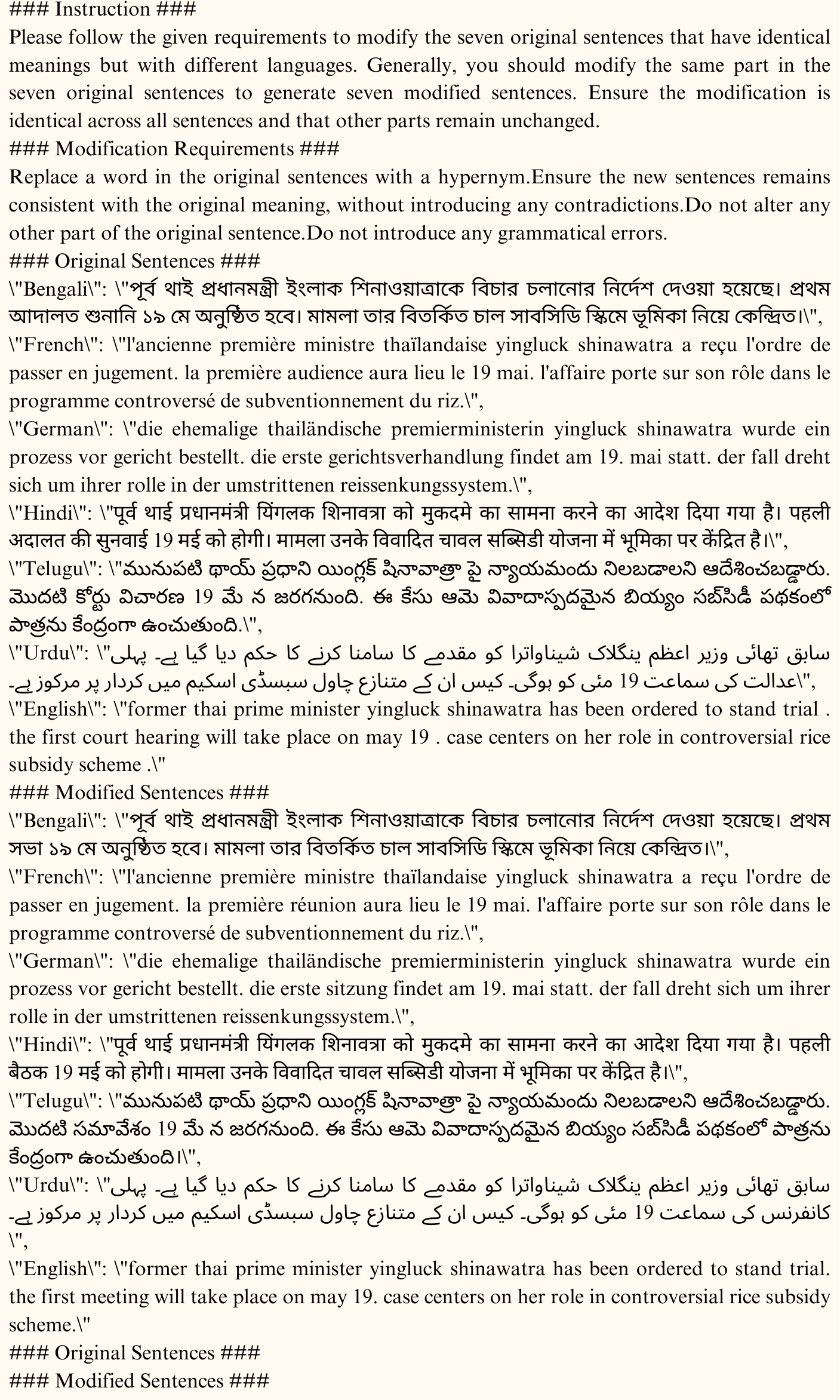} 
    \caption{The prompt designed to direct GPT-4o in generating perturbation texts for the Hypernym.}
    \label{fig:Hypernym}
\end{figure}

\begin{figure}[h!]
    \centering
    \includegraphics[width=\columnwidth]{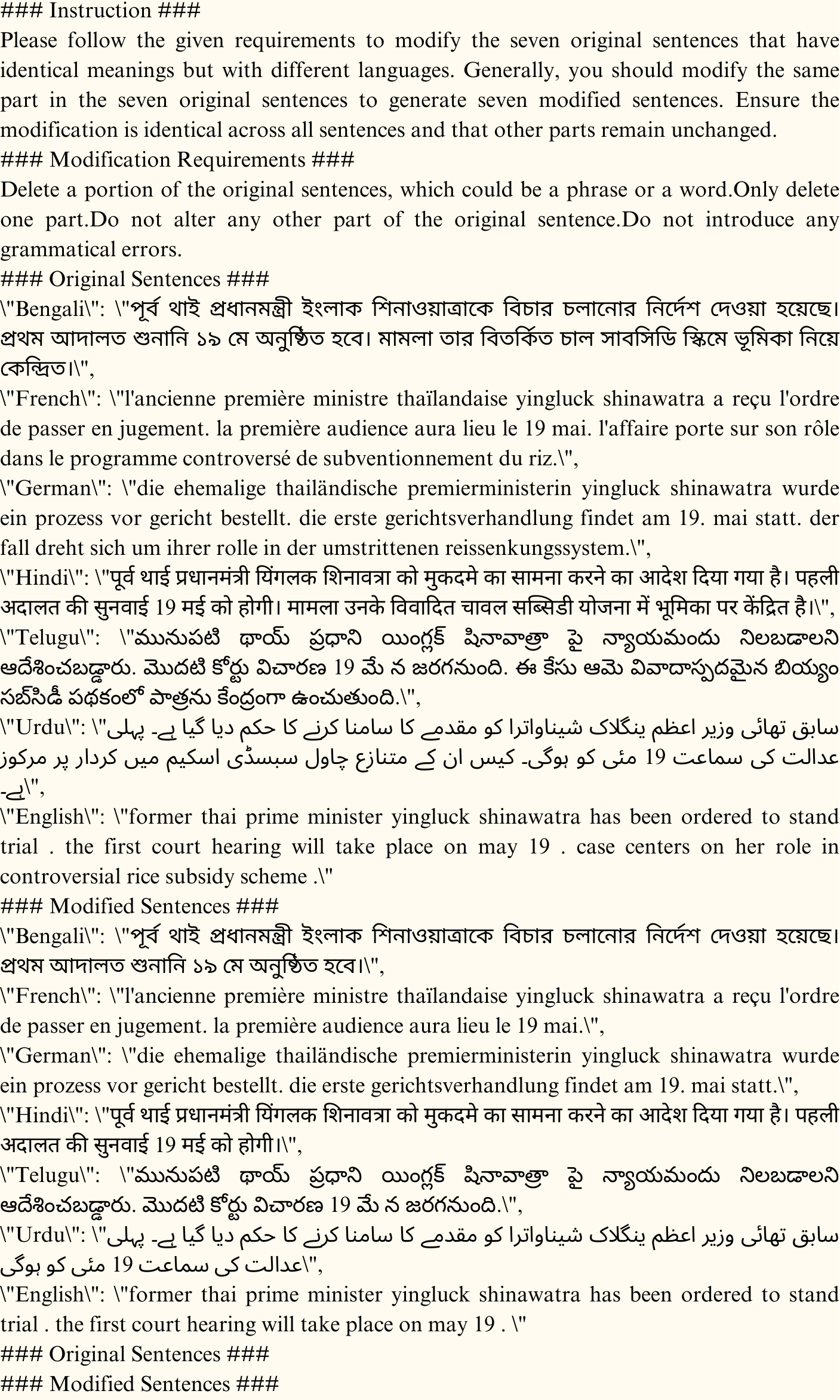} 
    \caption{The prompt designed to direct GPT-4o in generating perturbation texts for the Sentence Deletion.}
    \label{fig:Sentence Deletion}
\end{figure}

\begin{figure}[h!]
    \centering
    \includegraphics[width=\columnwidth]{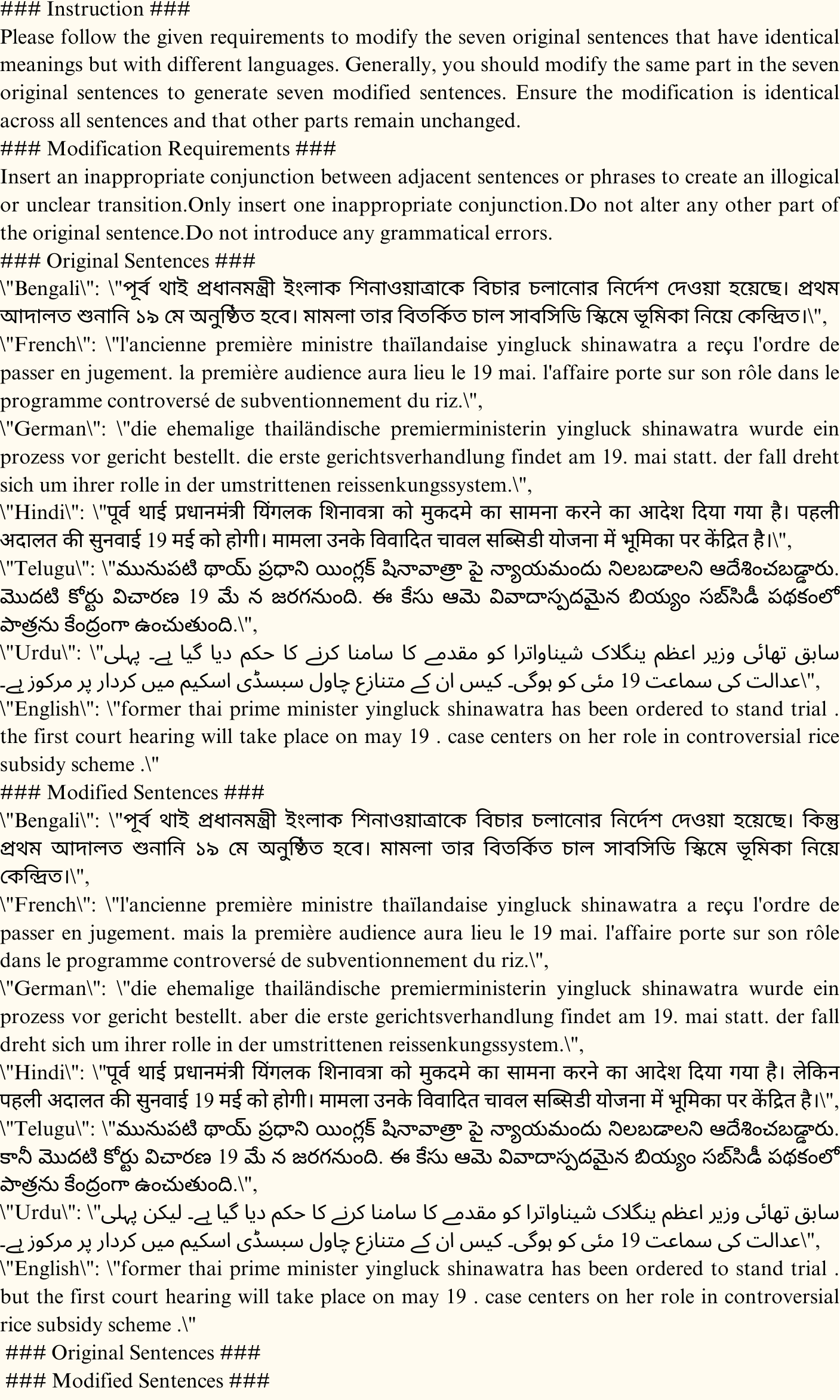} 
    \caption{The prompt designed to direct GPT-4o in generating perturbation texts for the Improper Connective.}
    \label{fig:Improper Connective}
\end{figure}

\begin{figure}[h!]
    \centering
    \includegraphics[width=\columnwidth]{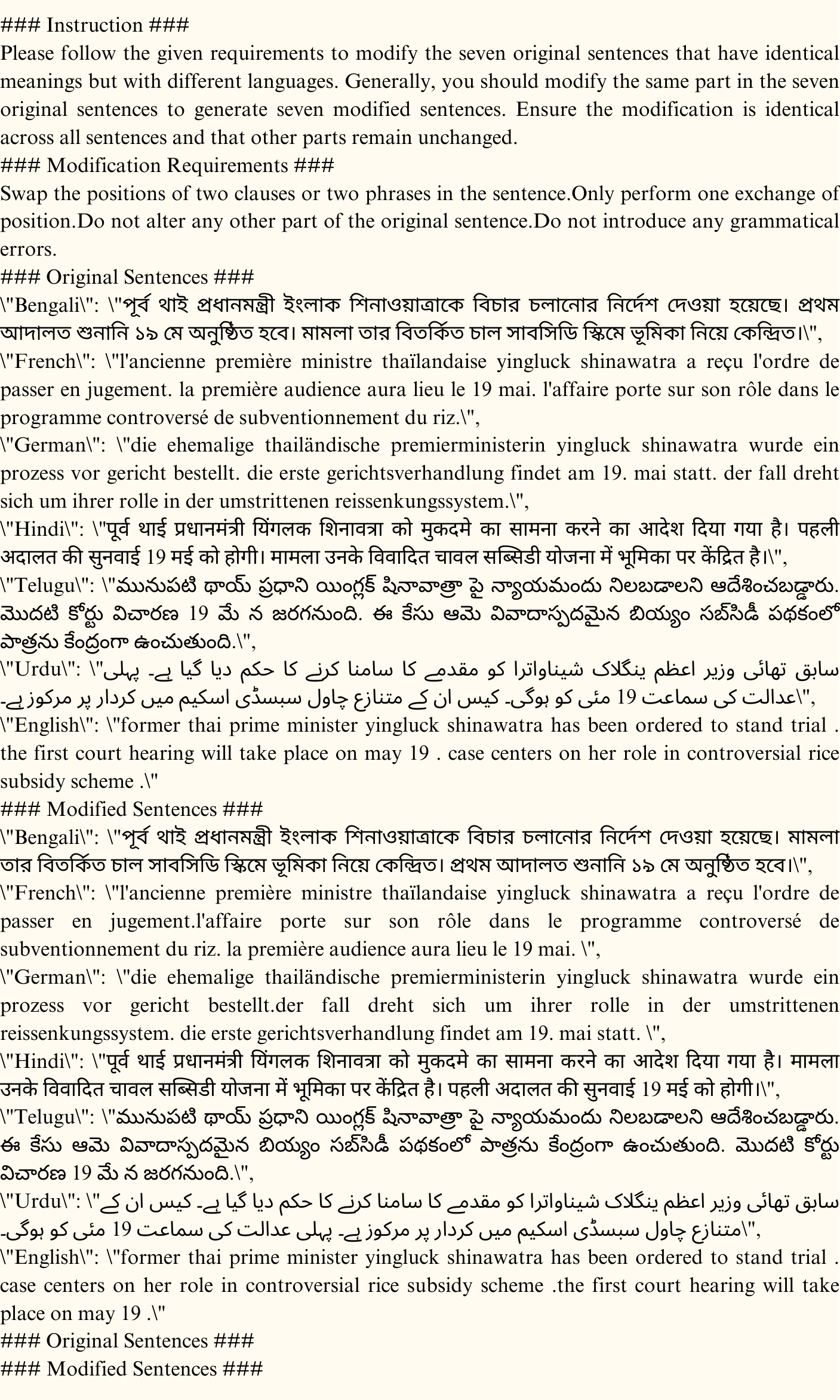} 
    \caption{The prompt designed to direct GPT-4o in generating perturbation texts for the Sentence Exchange.}
    \label{fig:Sentence Exchange}
\end{figure}

\begin{figure}[h!]
    \centering
    \includegraphics[width=\columnwidth]{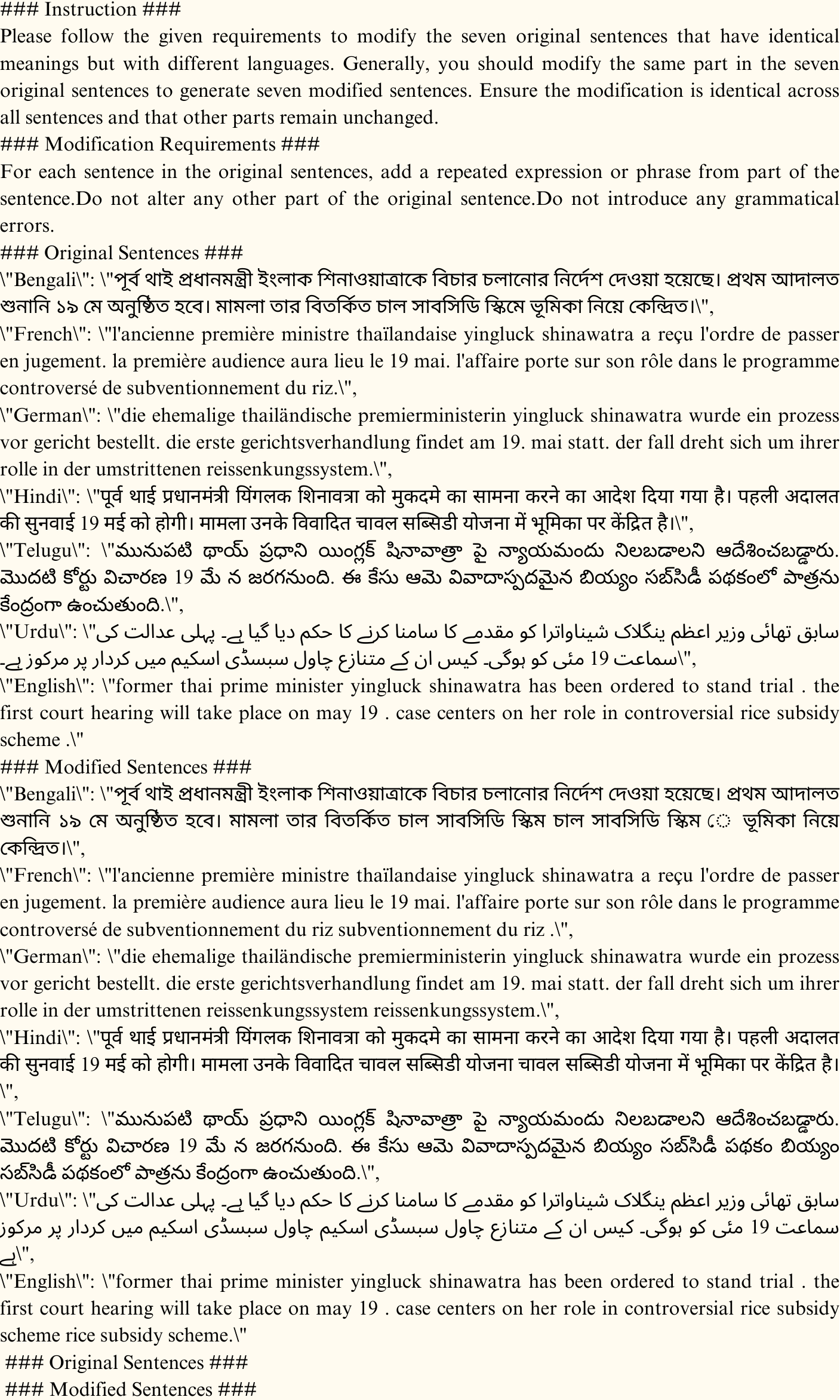} 
    \caption{The prompt designed to direct GPT-4o in generating perturbation texts for the Repetition.}
    \label{fig:Repetition}
\end{figure}

\begin{figure}[h!]
    \centering
    \includegraphics[width=\columnwidth]{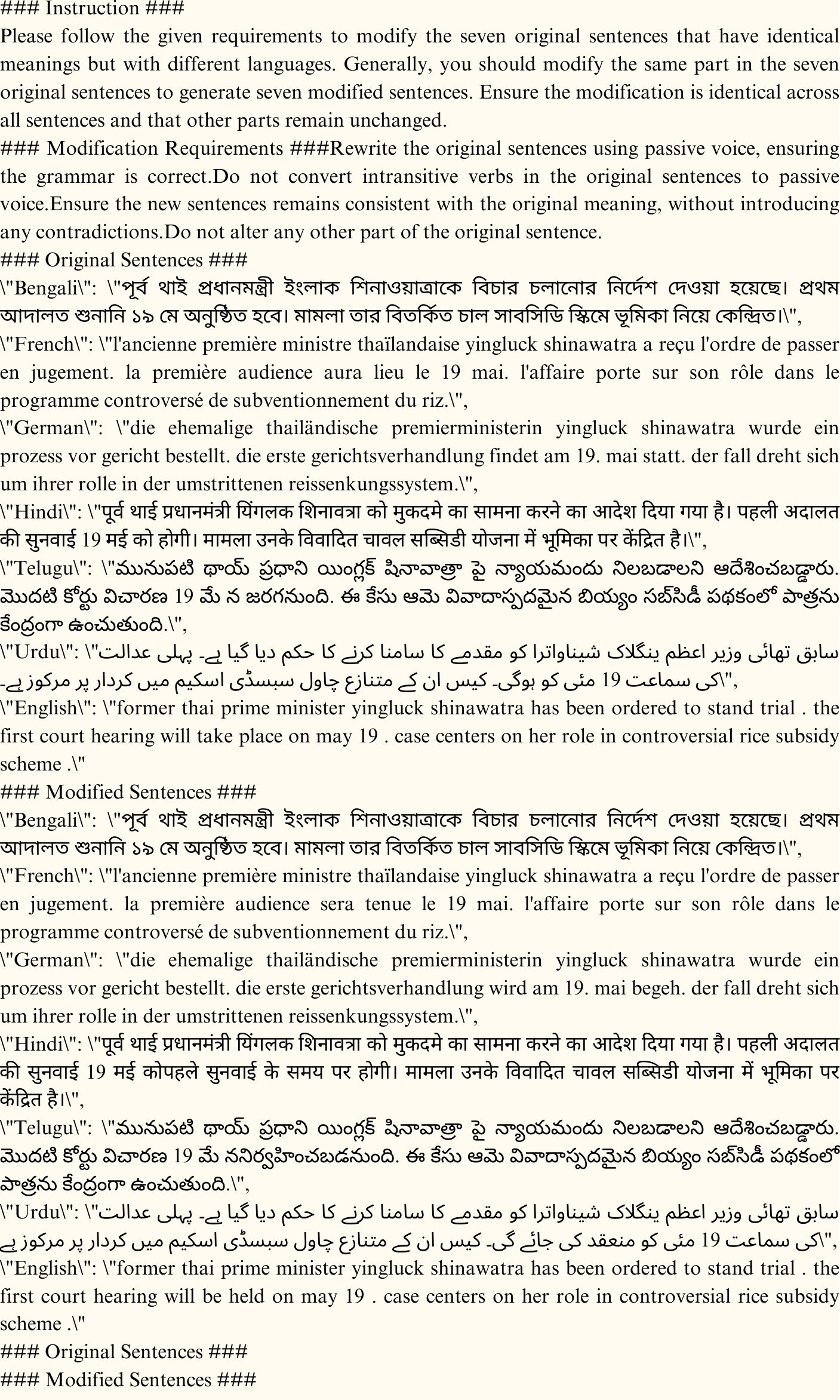} 
    \caption{The prompt designed to direct GPT-4o in generating perturbation texts for the Passive Voice.}
    \label{fig:Passive Voice}
\end{figure}

The prompt utilized to generate perturbation texts for the Uncommon Phrase is depicted in \autoref{fig:不常见的短语prompt}. An illustration of the prompt employed to generate perturbation texts for the Complex Sentence is provided in \autoref{fig:Complex Sentence}. In a similar vein, the prompt used for generating perturbation texts for the Complement is showcased in \autoref{fig:Complement}. The prompt for the Continuation is highlighted in \autoref{fig:Continuation}, whereas the corresponding prompt for the Different Entity is presented in \autoref{fig:Different Entity}. The prompt used for generating perturbation texts for the Negation is offered in \autoref{fig:Negation}, and the prompt for the Hypernym is outlined in \autoref{fig:Hypernym}. Furthermore, the prompt utilized for the Sentence Deletion is exemplified in \autoref{fig:Sentence Deletion}, with the prompt for the Improper Connective demonstrated in \autoref{fig:Improper Connective}. The prompt for generating perturbation texts for the Sentence Exchange is illustrated in \autoref{fig:Sentence Exchange}, while the prompt for the Repetition is conveyed in \autoref{fig:Repetition}. Lastly, the prompt used to generate perturbation texts for the Passive Voice is detailed in \autoref{fig:Passive Voice}.

\section{The prompts and evaluation criteria employed for the title generation and summary generation tasks.}
\label{sec:The prompts and evaluation criteria employed for the title generation and abstract generation tasks.}
\begin{figure}[h!]
    \centering
    \includegraphics[width=\columnwidth]{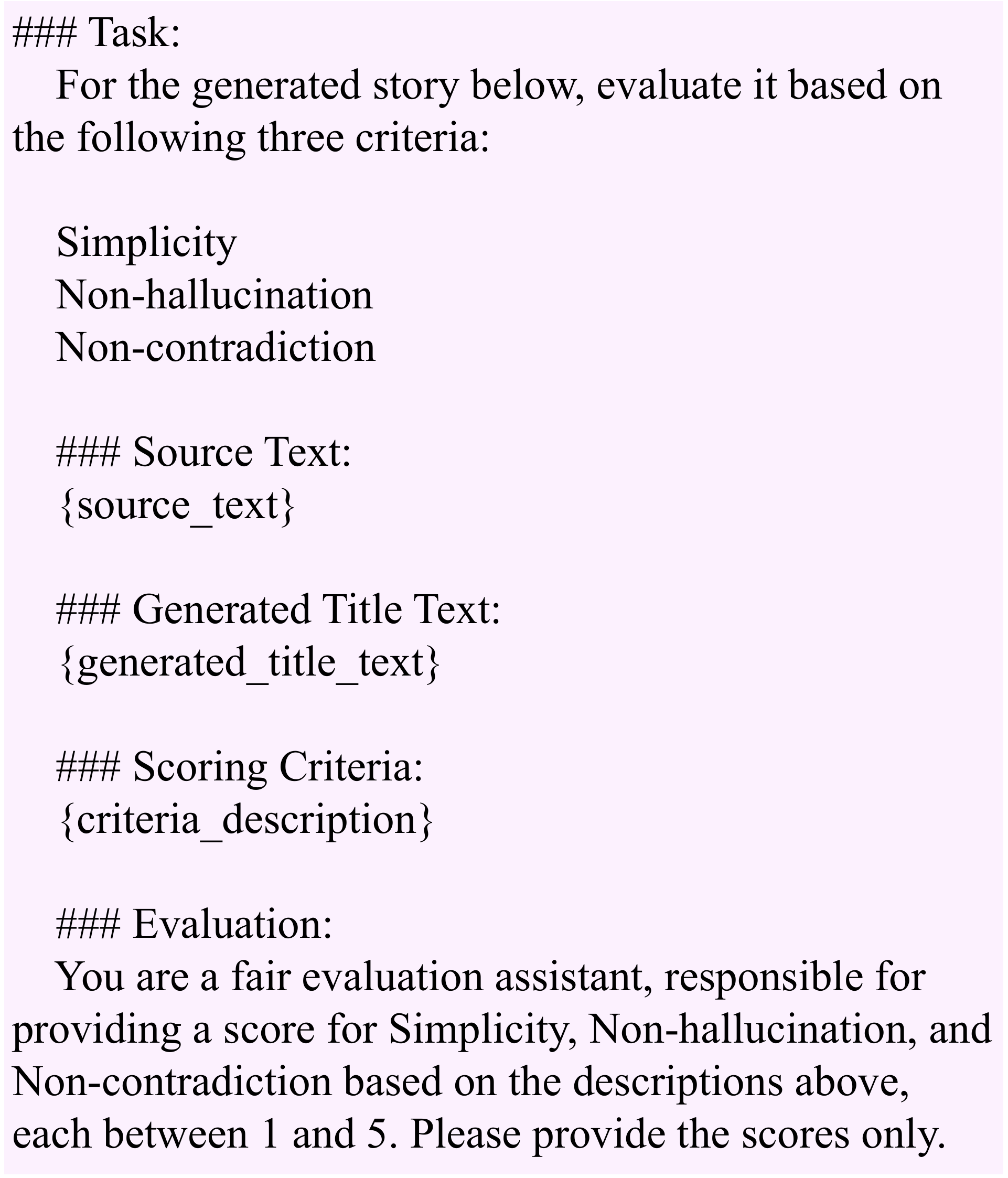} 
    \caption{Prompts for evaluating the task of title generation.}
    \label{fig:标题生成任务评价prompt}
\end{figure}

\begin{figure}[h!]
    \centering
    \includegraphics[width=\columnwidth]{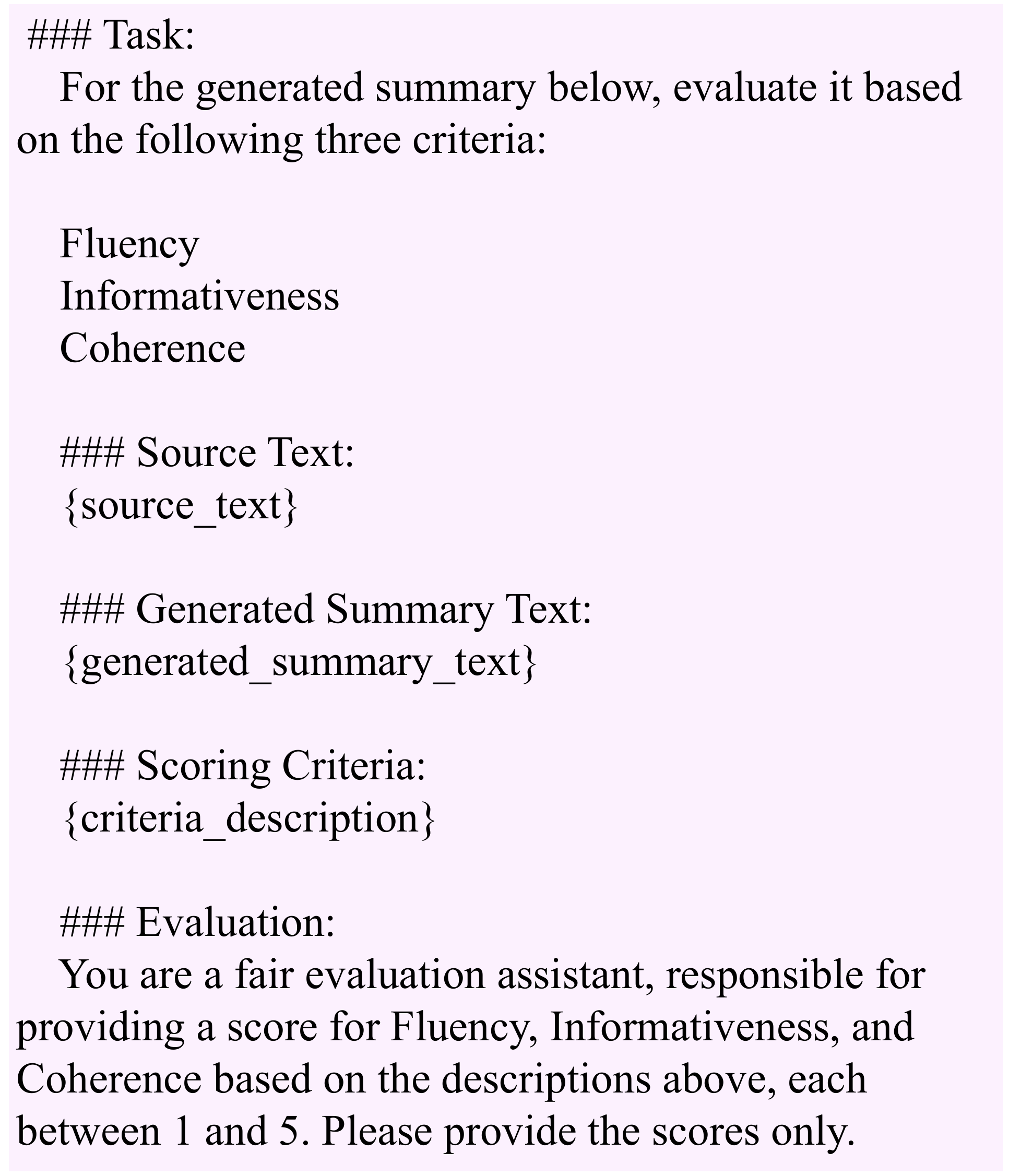} 
    \caption{Prompts for evaluating the task of summary generation.}
    \label{fig:摘要生成任务评价prompt}
\end{figure}

\begin{figure}[h!]
    \centering
    \includegraphics[width=\columnwidth]{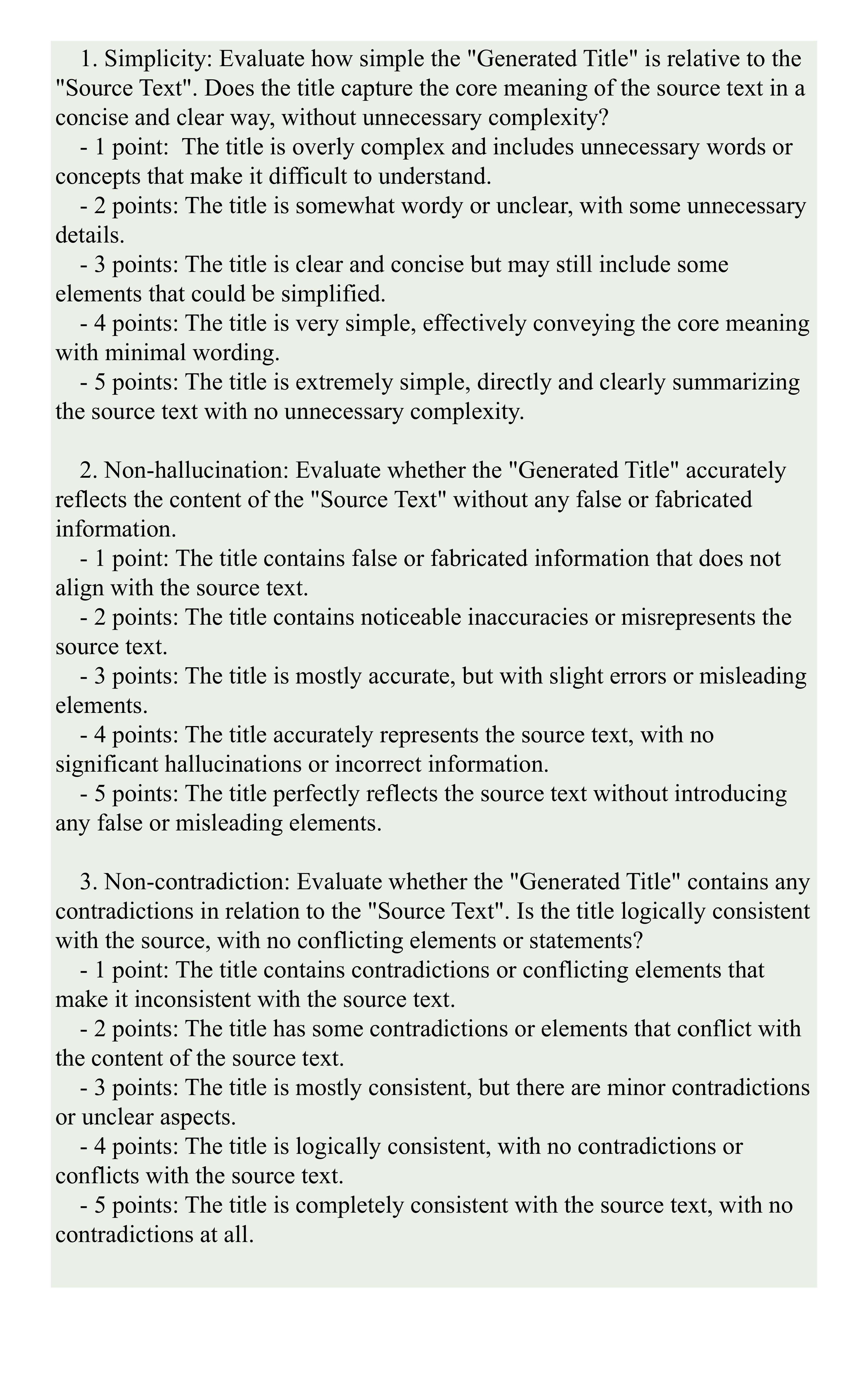} 
    \caption{Evaluation criteria for evaluating the task of title generation.}
    \label{fig:标题生成任务评价准测}
\end{figure}

\begin{figure}[h!]
    \centering
    \includegraphics[width=\columnwidth]{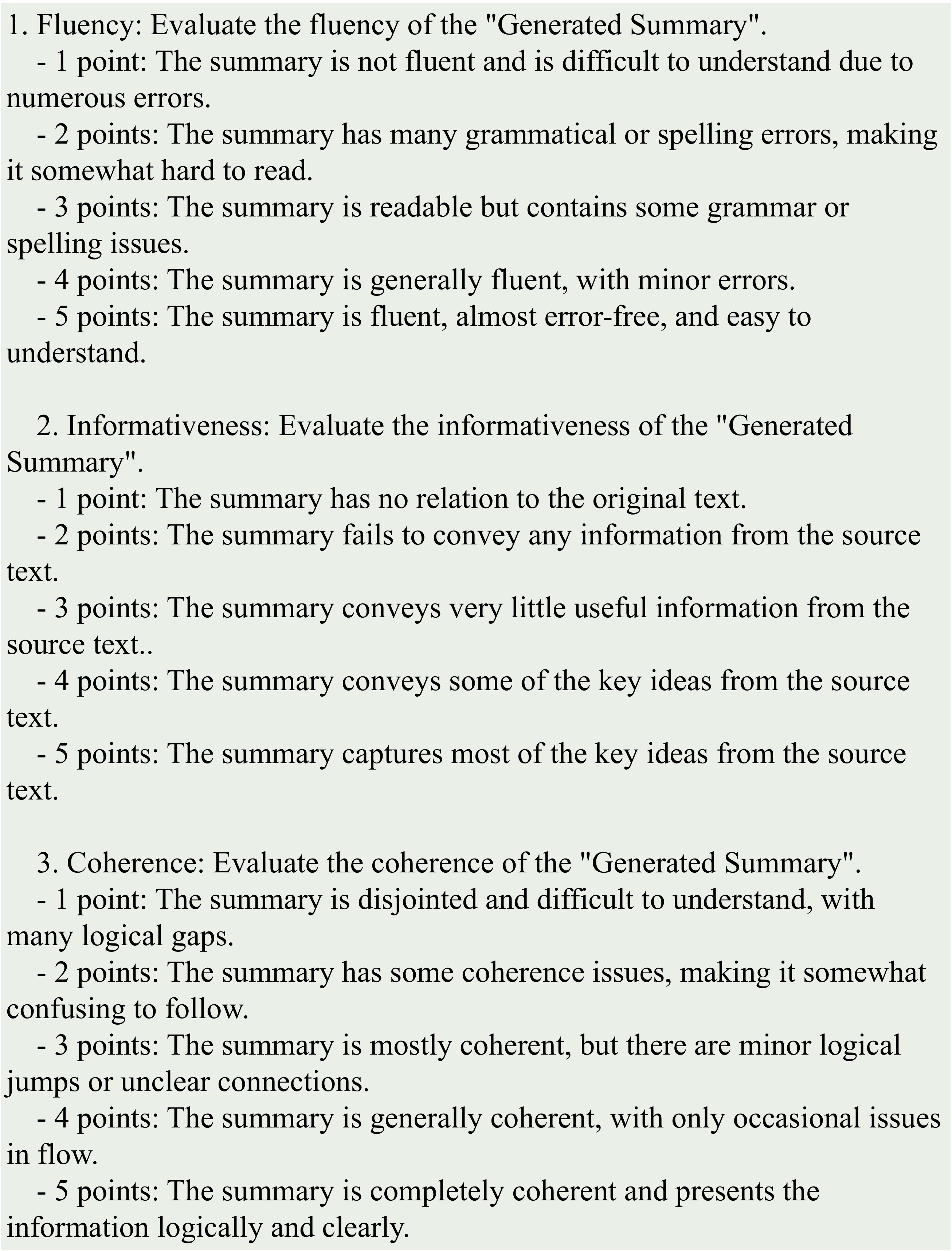} 
    \caption{Evaluation criteria for evaluating the task of summary generation.}
    \label{fig:摘要评价准则}
\end{figure}

This section offers a comprehensive delineation of the prompts and evaluation criteria utilized in the assessment of the title generation and summary generation tasks. The prompts utilized for the title generation task are presented in \autoref{fig:标题生成任务评价prompt}, while \autoref{fig:摘要生成任务评价prompt} illustrates the prompts applied to the summary generation task. The evaluation criteria for the title generation task are delineated in \autoref{fig:标题生成任务评价准测}, and \autoref{fig:摘要评价准则} outlines the evaluation criteria for the abstract generation task.


\section{The significance test results of the multilingual evaluation ability correlations for partial LLMs}
\label{sec:Detailed Significance Testing and Results for Scenarios with Reference Answers}
The significance test results for the cross-linguistic evaluation capabilities of Qwen2.5-72B and Llama-3.1-70B-Instruct across different language pairs are presented in \autoref{tab:qwen2.5_significance} and \autoref{tab:Llama-3.1-70B-Instruct_significance}. A p-value below 0.05 indicates that the model with a larger parameter size exhibits significantly higher evaluation performance in the first language of the pair compared to the second. Conversely, a p-value equal to or above 0.05 suggests that the model’s evaluation performance in the first language is less than or not significantly different from that in the second language. The null and alternative hypotheses are provided below.

\begin{align*}
H_0\colon \rho\left( X,Z\right) &\leq \rho\left( Y,Z\right) \\
H_1\colon \rho\left( X,Z\right) &> \rho\left( Y,Z\right)
\end{align*}
where $X$ corresponds to the evaluation score of a large-scale parameter model applied to a specific language, and $Y$ represents the evaluation score of the same model applied to a different language. $Z$ denotes the gold score. The Pearson correlation coefficient between the model's performance on the first language and the gold score is quantified as $\rho\left( X,Z\right)$. Correspondingly, the Pearson correlation coefficient between the model's performance on the second language and the gold score is expressed as $\rho\left( Y,Z\right)$.

\begin{table*} 
\small
\centering 
\begin{tabular}{cccc}
\toprule
{\color[HTML]{333333} \textbf{Language pairs}}    & {\color[HTML]{333333} \textbf{p-values}} & {\color[HTML]{333333} \textbf{Language pairs}}  & {\color[HTML]{333333} \textbf{p-values}} \\ \midrule
{\color[HTML]{333333} French vs Bengali}          & {\color[HTML]{333333} 0.000}            & {\color[HTML]{333333} German vs Hindi}         & {\color[HTML]{333333} 0.000}            \\
German vs Bengali                                 & 0.000                                  & German vs Telugu                               & 0.000                                  \\
Hindi vs Bengali                                  & 0.931                                  & German vs Urdu                                 & 0.000                                  \\
Bengali vs Telugu                                 & 0.002                                  & En vs German                                   & 0.009                                  \\
Urdu vs Bengali                                   & 0.745                                  & Hindi vs Telugu                                & 0.030                                  \\
En vs Bengali                                     & 0.000                                  & Urdu vs Hindi                                  & 0.206                                  \\
German vs French                                  & 0.197                                  & En vs Hindi                                    & 0.000                                  \\
French vs Hindi                                   & 0.000                                  & Urdu vs Telugu                                 & 0.006                                  \\
French vs Telugu                                  & 0.000                                  & En vs Telugu                                   & 0.000                                  \\
French vs Urdu                                    & 0.000                                  & En vs Urdu                                     & 0.000                                  \\
En vs French                                      & 0.000                                  &                                                &                                        \\ \bottomrule
\end{tabular}
\caption{The results of the significance testing for the cross-linguistic evaluation performance of Qwen2.5-72B.} 
\label{tab:qwen2.5_significance} 
\end{table*}

\begin{table*} 
\small
\centering 
\begin{tabular}{cccc}
\toprule
{\color[HTML]{333333} \textbf{Language pairs}}    & {\color[HTML]{333333} \textbf{p-values}} & {\color[HTML]{333333} \textbf{Language pairs}}  & {\color[HTML]{333333} \textbf{p-values}} \\ \midrule
{\color[HTML]{333333} French vs Bengali}          & {\color[HTML]{333333} 0.000}            & {\color[HTML]{333333} German vs Hindi}         & {\color[HTML]{333333} 0.014}            \\
German vs Bengali                                 & 0.000                                  & German vs Telugu                               & 0.000                                  \\
Hindi vs Bengali                                  & 0.000                                  & Urdu vs German                                 & 0.395                                  \\
Bengali vs Telugu                                 & 0.018                                  & En vs German                                   & 0.000                                  \\
Urdu vs Bengali                                   & 0.000                                  & Hindi vs Telugu                                & 0.000                                  \\
En vs Bengali                                     & 0.000                                  & Urdu vs Hindi                                  & 0.005                                  \\
German vs French                                  & 0.104                                  & En vs Hindi                                    & 0.000                                  \\
French vs Hindi                                   & 0.168                                  & Urdu vs Telugu                                 & 0.000                                  \\
French vs Telugu                                  & 0.000                                  & En vs Telugu                                   & 0.000                                  \\
Urdu vs French                                    & 0.070                                  & En vs Urdu                                     & 0.000                                  \\
En vs French                                      & 0.000                                  &                                                &                                        \\ \bottomrule
\end{tabular}
\caption{The results of the significance testing for the cross-linguistic evaluation performance of Llama-3.1-70B-Instruct.} 
\label{tab:Llama-3.1-70B-Instruct_significance} 
\end{table*}

\section{The sensitivity of the remaining LLMs to perturbation texts is presented.}
\label{sec:The sensitivity of the remaining nine LLMs to perturbation texts is presented.}

This section provides a thorough examination of the sensitivity of the remaining LLMs to perturbation texts in both title and abstract generation tasks across seven languages. Specifically, \autoref{tab:gemma-2-9b-it-title} and \autoref{tab:The sensitivity of gemma-2-9b-it to perturbation texts in the summary generation task} delineate gemma-2-9b-it's sensitivity in these tasks. \autoref{tab:glm-4-9B-chat-title} and \autoref{tab:The sensitivity of glm-4-9B-chat to perturbation texts in the summary generation task} elaborate on the sensitivity of glm-4-9B-chat. Furthermore, \autoref{tab:glm4-flash-title} and \autoref{tab:glm4-flash summary generation task} scrutinize the sensitivity of glm4-flash. \autoref{tab:Qwen2.5-72b-title} and \autoref{tab:Qwen2.5-72B summary generation task} emphasize the sensitivity of Qwen2.5-72B, with \autoref{tab:Llama-3.1-70B-Instruct-title} and \autoref{tab:Llama-3.1-70B-Instruct summary generation task} revisiting Llama-3.1-70B-Instruct. Ultimately, \autoref{tab: Qwen-turbo-title} and \autoref{tab:Qwen-turbo summary generation task} elucidate the sensitivity of Qwen-Turbo across these tasks and languages. In the tables below, blue highlights the language demonstrating the highest sensitivity to each perturbation type, while orange denotes the language with the lowest sensitivity for the respective perturbations.

\begin{table*} 
\small
\centering 
\begin{tabular}{llllllll}
\toprule
\multicolumn{8}{c}{{\color[HTML]{000000} \textbf{Title Generation Task}}} \\ \cmidrule(lr){2-8}
\multicolumn{1}{c}{{\color[HTML]{000000} \textbf{Perturbations}}} & \multicolumn{1}{c}{{\color[HTML]{000000} \textbf{Bengali}}} & \multicolumn{1}{c}{{\color[HTML]{000000} \textbf{French}}} & \multicolumn{1}{c}{{\color[HTML]{000000} \textbf{German}}} & \multicolumn{1}{c}{{\color[HTML]{000000} \textbf{Hindi}}} & \multicolumn{1}{c}{{\color[HTML]{000000} \textbf{Telugu}}} & \multicolumn{1}{c}{{\color[HTML]{000000} \textbf{Urdu}}} & \multicolumn{1}{c}{{\color[HTML]{000000} \textbf{English}}} \\ \midrule
{\color[HTML]{000000} Uncommon Phrase(Sim)}                       & {\color[HTML]{000000} 0.225}                                & {\color[HTML]{000000} 0.285}                               & {\color[HTML]{000000} 0.330}                               & {\color[HTML]{34CDF9} 0.545}                              & {\color[HTML]{000000} 0.220}                               & {\color[HTML]{F56B00} -0.100}                            & {\color[HTML]{000000} 0.375}                                \\
{\color[HTML]{000000} Complex Sentence(Sim)}                      & {\color[HTML]{000000} 0.155}                                & {\color[HTML]{000000} 0.080}                               & {\color[HTML]{000000} 0.190}                               & {\color[HTML]{000000} 0.000}                              & {\color[HTML]{000000} 0.095}                               & {\color[HTML]{34CDF9} 0.360}                             & {\color[HTML]{F56B00} -0.025}                               \\
{\color[HTML]{000000} Complement(Non-hal)}                        & {\color[HTML]{000000} 0.340}                                & {\color[HTML]{000000} 0.330}                               & {\color[HTML]{000000} 0.340}                               & {\color[HTML]{34CDF9} 0.355}                              & {\color[HTML]{000000} 0.330}                               & {\color[HTML]{F56B00} 0.165}                             & {\color[HTML]{000000} 0.300}                                \\
{\color[HTML]{000000} Continuation(Non-hal)}                      & {\color[HTML]{000000} 0.340}                                & {\color[HTML]{000000} 0.335}                               & {\color[HTML]{000000} 0.340}                               & {\color[HTML]{F56B00} 0.270}                              & {\color[HTML]{000000} 0.310}                               & {\color[HTML]{34CDF9} 0.740}                             & {\color[HTML]{000000} 0.275}                                \\
{\color[HTML]{000000} Different Entity(Non-con)}                  & {\color[HTML]{000000} 1.050}                                & {\color[HTML]{000000} 0.860}                               & {\color[HTML]{000000} 1.050}                               & {\color[HTML]{000000} 1.005}                              & {\color[HTML]{F56B00} 0.795}                               & {\color[HTML]{34CDF9} 1.340}                             & {\color[HTML]{000000} 1.235}                                \\
{\color[HTML]{000000} Negation(Non-con)}                          & {\color[HTML]{000000} 1.105}                                & {\color[HTML]{000000} 1.120}                               & {\color[HTML]{000000} 1.105}                               & {\color[HTML]{000000} 1.015}                              & {\color[HTML]{000000} 1.490}                               & {\color[HTML]{F56B00} -0.025}                            & {\color[HTML]{34CDF9} 1.875}                                \\
{\color[HTML]{000000} Average}                                    & 0.536                                                         & 0.502                                                         & 0.559                                                         & 0.532                                                   & 0.540                                                 & 0.413                                                & 0.673                                                 \\ \bottomrule
\end{tabular}
\caption{The sensitivity of gemma-2-9b-it to perturbation texts in the title generation task across seven languages.} 
\label{tab:gemma-2-9b-it-title} 
\end{table*}

\begin{table*} 
\small
\centering 
\begin{tabular}{llllllll} 
\toprule
\multicolumn{8}{c}{{\color[HTML]{000000} \textbf{Summary Generation Task}}} \\ \cmidrule(lr){2-8}
\multicolumn{1}{c}{{\color[HTML]{000000} \textbf{Perturbations}}} & \multicolumn{1}{c}{{\color[HTML]{000000} \textbf{Bengali}}} & \multicolumn{1}{c}{{\color[HTML]{000000} \textbf{French}}} & \multicolumn{1}{c}{{\color[HTML]{000000} \textbf{German}}} & \multicolumn{1}{c}{{\color[HTML]{000000} \textbf{Hindi}}} & \multicolumn{1}{c}{{\color[HTML]{000000} \textbf{Telugu}}} & \multicolumn{1}{c}{{\color[HTML]{000000} \textbf{Urdu}}} & \multicolumn{1}{c}{{\color[HTML]{000000} \textbf{English}}} \\ \midrule
{\color[HTML]{000000} Hypernym(Inf)}                              & {\color[HTML]{000000} -0.065}                               & {\color[HTML]{34CDF9} 0.010}                               & {\color[HTML]{000000} -0.070}                              & {\color[HTML]{F56B00} -0.075}                             & {\color[HTML]{F56B00} -0.075}                              & {\color[HTML]{34CDF9} 0.010}                             & {\color[HTML]{000000} -0.035}                               \\
{\color[HTML]{000000} Sentence Deletion(Inf)}                     & {\color[HTML]{000000} 0.065}                                & {\color[HTML]{34CDF9} 0.130}                               & {\color[HTML]{F56B00} 0.035}                               & {\color[HTML]{000000} 0.040}                              & {\color[HTML]{000000} 0.050}                               & {\color[HTML]{000000} 0.115}                             & {\color[HTML]{F56B00} 0.035}                                \\
{\color[HTML]{000000} Improper Connective(Coh)}                   & {\color[HTML]{F56B00} 0.060}                                & {\color[HTML]{000000} 0.160}                               & {\color[HTML]{000000} 0.180}                               & {\color[HTML]{000000} 0.140}                              & {\color[HTML]{000000} 0.140}                               & {\color[HTML]{000000} 0.130}                             & {\color[HTML]{34CDF9} 0.185}                                \\
{\color[HTML]{000000} Sentence Exchange(Coh)}                     & {\color[HTML]{000000} 0.130}                                & {\color[HTML]{000000} 0.090}                               & {\color[HTML]{000000} 0.195}                               & {\color[HTML]{000000} 0.060}                              & {\color[HTML]{000000} 0.060}                               & {\color[HTML]{F56B00} 0.050}                             & {\color[HTML]{34CDF9} 0.210}                                \\
{\color[HTML]{000000} Repetition(Flu)}                            & {\color[HTML]{000000} 0.245}                                & {\color[HTML]{000000} 0.255}                               & {\color[HTML]{34CDF9} 0.300}                               & {\color[HTML]{000000} 0.125}                              & {\color[HTML]{F56B00} 0.105}                               & {\color[HTML]{000000} 0.225}                             & {\color[HTML]{34CDF9} 0.300}                                \\
{\color[HTML]{000000} Passive Voice(Flu)}                         & {\color[HTML]{F56B00} -0.065}                               & {\color[HTML]{000000} 0.030}                               & {\color[HTML]{000000} 0.390}                               & {\color[HTML]{000000} 0.035}                              & {\color[HTML]{000000} 0.025}                               & {\color[HTML]{000000} 0.025}                             & {\color[HTML]{34CDF9} 0.430}                                \\
{\color[HTML]{000000} Average}                                    & {\color[HTML]{060607} 0.062}                                & {\color[HTML]{060607} 0.113}                               & {\color[HTML]{060607} 0.172}                               & {\color[HTML]{060607} 0.054}                              & {\color[HTML]{000000} 0.053}                               & {\color[HTML]{060607} 0.093}                             & {\color[HTML]{060607} 0.188}                                \\ \bottomrule
\end{tabular}
\caption{The sensitivity of gemma-2-9b-it to perturbation texts in the summary generation task across seven languages.} 
\label{tab:The sensitivity of gemma-2-9b-it to perturbation texts in the summary generation task}
\end{table*}

\begin{table*} 
\small
\centering 
\begin{tabular}{llllllll}
\toprule
\multicolumn{8}{c}{{\color[HTML]{000000} \textbf{Title Generation Task}}} \\ \cmidrule(lr){2-8}
\multicolumn{1}{c}{{\color[HTML]{000000} \textbf{Perturbations}}} & \multicolumn{1}{c}{{\color[HTML]{000000} \textbf{Bengali}}} & \multicolumn{1}{c}{{\color[HTML]{000000} \textbf{French}}} & \multicolumn{1}{c}{{\color[HTML]{000000} \textbf{German}}} & \multicolumn{1}{c}{{\color[HTML]{000000} \textbf{Hindi}}} & \multicolumn{1}{c}{{\color[HTML]{000000} \textbf{Telugu}}} & \multicolumn{1}{c}{{\color[HTML]{000000} \textbf{Urdu}}} & \multicolumn{1}{c}{{\color[HTML]{000000} \textbf{English}}} \\ \midrule
{\color[HTML]{000000} Uncommon Phrase(Sim)}                       & {\color[HTML]{000000} 0.105}                                & {\color[HTML]{34CDF9} 0.130}                               & {\color[HTML]{F56B00} 0.045}                               & {\color[HTML]{000000} 0.095}                              & {\color[HTML]{000000} 0.100}                               & {\color[HTML]{000000} 0.085}                             & {\color[HTML]{000000} 0.090}                                \\
{\color[HTML]{000000} Complex Sentence(Sim)}                      & {\color[HTML]{000000} 0.005}                                & {\color[HTML]{000000} 0.050}                               & {\color[HTML]{000000} 0.035}                               & {\color[HTML]{000000} 0.100}                              & {\color[HTML]{34CDF9} 0.140}                               & {\color[HTML]{F56B00} 0.000}                             & {\color[HTML]{F56B00} 0.000}                                \\
{\color[HTML]{000000} Complement(Non-hal)}                        & {\color[HTML]{000000} 0.115}                                & {\color[HTML]{000000} 0.105}                               & {\color[HTML]{000000} 0.115}                               & {\color[HTML]{34CDF9} 0.140}                              & {\color[HTML]{000000} 0.100}                               & {\color[HTML]{F56B00} 0.090}                             & {\color[HTML]{F56B00} 0.090}                                \\
{\color[HTML]{000000} Continuation(Non-hal)}                      & {\color[HTML]{000000} 0.040}                                & {\color[HTML]{34CDF9} 0.185}                               & {\color[HTML]{000000} 0.145}                               & {\color[HTML]{000000} 0.115}                              & {\color[HTML]{F56B00} 0.025}                               & {\color[HTML]{000000} 0.065}                             & {\color[HTML]{000000} 0.085}                                \\
{\color[HTML]{000000} Different Entity(Non-con)}                  & {\color[HTML]{000000} 0.170}                                & {\color[HTML]{000000} 0.245}                               & {\color[HTML]{000000} 0.260}                               & {\color[HTML]{000000} 0.200}                              & {\color[HTML]{F56B00} 0.140}                               & {\color[HTML]{000000} 0.210}                             & {\color[HTML]{34CDF9} 0.275}                                \\
{\color[HTML]{000000} Negation(Non-con)}                          & {\color[HTML]{000000} 0.400}                                & {\color[HTML]{000000} 0.410}                               & {\color[HTML]{000000} 0.335}                               & {\color[HTML]{000000} 0.460}                              & {\color[HTML]{F56B00} 0.160}                               & {\color[HTML]{000000} 0.450}                             & {\color[HTML]{34CDF9} 0.565}                                \\
{\color[HTML]{000000} Average}                                    & 0.139                                                         & 0.188                                                         & 0.156                                                         & 0.185                                                   & 0.111                                                 & 0.150                                                & 0.184                                                 \\ \bottomrule
\end{tabular}
\caption{The sensitivity of glm-4-9B-chat to perturbation texts in the title generation task across seven languages.} 
\label{tab:glm-4-9B-chat-title} 
\end{table*}

\begin{table*} 
\small
\centering 
\begin{tabular}{llllllll}
\toprule
\multicolumn{8}{c}{{\color[HTML]{000000} \textbf{Summary Generation Task}}} \\ \cmidrule(lr){2-8}
\multicolumn{1}{c}{{\color[HTML]{000000} \textbf{Perturbations}}} & \multicolumn{1}{c}{{\color[HTML]{000000} \textbf{Bengali}}} & \multicolumn{1}{c}{{\color[HTML]{000000} \textbf{French}}} & \multicolumn{1}{c}{{\color[HTML]{000000} \textbf{German}}} & \multicolumn{1}{c}{{\color[HTML]{000000} \textbf{Hindi}}} & \multicolumn{1}{c}{{\color[HTML]{000000} \textbf{Telugu}}} & \multicolumn{1}{c}{{\color[HTML]{000000} \textbf{Urdu}}} & \multicolumn{1}{c}{{\color[HTML]{000000} \textbf{English}}} \\ \midrule
{\color[HTML]{000000} Hypernym(Inf)}                              & {\color[HTML]{000000} 0.005}                                & {\color[HTML]{000000} 0.020}                               & {\color[HTML]{F56B00} -0.115}                              & {\color[HTML]{000000} 0.015}                              & {\color[HTML]{000000} -0.010}                              & {\color[HTML]{000000} 0.020}                             & {\color[HTML]{34CDF9} 0.075}                                \\
{\color[HTML]{000000} Sentence Deletion(Inf)}                     & {\color[HTML]{F56B00} 0.025}                                & {\color[HTML]{34CDF9} 0.135}                               & {\color[HTML]{000000} 0.165}                               & {\color[HTML]{000000} 0.085}                              & {\color[HTML]{000000} 0.050}                               & {\color[HTML]{000000} 0.115}                             & {\color[HTML]{000000} 0.075}                                \\
{\color[HTML]{000000} Improper Connective(Coh)}                   & {\color[HTML]{000000} 0.105}                                & {\color[HTML]{000000} 0.095}                               & {\color[HTML]{000000} 0.035}                               & {\color[HTML]{000000} 0.020}                              & {\color[HTML]{F56B00} -0.045}                              & {\color[HTML]{34CDF9} 0.110}                             & {\color[HTML]{000000} 0.090}                                \\
{\color[HTML]{000000} Sentence Exchange(Coh)}                     & {\color[HTML]{000000} 0.070}                                & {\color[HTML]{000000} 0.100}                               & {\color[HTML]{000000} 0.070}                               & {\color[HTML]{34CDF9} 0.090}                              & {\color[HTML]{F56B00} -0.010}                              & {\color[HTML]{000000} 0.065}                             & {\color[HTML]{000000} 0.060}                                \\
{\color[HTML]{000000} Repetition(Flu)}                            & {\color[HTML]{F56B00} 0.010}                                & {\color[HTML]{000000} 0.120}                               & {\color[HTML]{000000} 0.155}                               & {\color[HTML]{000000} 0.110}                              & {\color[HTML]{000000} 0.140}                               & {\color[HTML]{000000} 0.230}                             & {\color[HTML]{34CDF9} 0.295}                                \\
{\color[HTML]{000000} Passive Voice(Flu)}                         & {\color[HTML]{000000} 0.015}                                & {\color[HTML]{F56B00} -0.040}                              & {\color[HTML]{000000} 0.005}                               & {\color[HTML]{000000} -0.015}                             & {\color[HTML]{000000} 0.005}                               & {\color[HTML]{000000} 0.005}                             & {\color[HTML]{34CDF9} 0.025}                                \\
{\color[HTML]{000000} Average}                                    & {\color[HTML]{060607} 0.038}                                & {\color[HTML]{060607} 0.072}                               & {\color[HTML]{060607} 0.053}                               & {\color[HTML]{060607} 0.051}                              & {\color[HTML]{000000} 0.022}                               & {\color[HTML]{000000} 0.091}                             & {\color[HTML]{000000} 0.103}                                \\ \bottomrule
\end{tabular}
\caption{The sensitivity of glm-4-9B-chat to perturbation texts in the summary generation task across seven languages.} 
\label{tab:The sensitivity of glm-4-9B-chat to perturbation texts in the summary generation task}
\end{table*}

\begin{table*} 
\small
\centering 
\begin{tabular}{lccccccc}
\toprule
\multicolumn{8}{c}{{\color[HTML]{000000} \textbf{Title Generation Task}}} \\ \cmidrule(lr){2-8}
\multicolumn{1}{c}{{\color[HTML]{000000} \textbf{Perturbations}}} & {\color[HTML]{000000} \textbf{Bengali}} & {\color[HTML]{000000} \textbf{French}} & {\color[HTML]{000000} \textbf{German}} & {\color[HTML]{000000} \textbf{Hindi}} & {\color[HTML]{000000} \textbf{Telugu}} & {\color[HTML]{000000} \textbf{Urdu}} & {\color[HTML]{000000} \textbf{English}} \\ \midrule
{\color[HTML]{000000} Uncommon Phrase(Sim)}                       & {\color[HTML]{000000} 0.070} & {\color[HTML]{000000} 0.150} & {\color[HTML]{000000} 0.180} & {\color[HTML]{000000} 0.085} & {\color[HTML]{000000} 0.065} & {\color[HTML]{F56B00} 0.055} & {\color[HTML]{34CDF9} 0.260} \\
{\color[HTML]{000000} Complex Sentence(Sim)}                      & {\color[HTML]{000000} 0.030} & {\color[HTML]{000000} 0.090} & {\color[HTML]{000000} 0.150} & {\color[HTML]{000000} 0.060} & {\color[HTML]{000000} 0.025} & {\color[HTML]{F56B00} 0.015} & {\color[HTML]{34CDF9} 0.150} \\
{\color[HTML]{000000} Complement(Non-hal)}                        & {\color[HTML]{F56B00} 0.085} & {\color[HTML]{000000} 0.210} & {\color[HTML]{000000} 0.210} & {\color[HTML]{000000} 0.150} & {\color[HTML]{000000} 0.115} & {\color[HTML]{000000} 0.130} & {\color[HTML]{34CDF9} 0.290} \\
{\color[HTML]{000000} Continuation(Non-hal)}                      & {\color[HTML]{000000} 0.125} & {\color[HTML]{000000} 0.215} & {\color[HTML]{000000} 0.200} & {\color[HTML]{000000} 0.090} & {\color[HTML]{F56B00} 0.070} & {\color[HTML]{000000} 0.165} & {\color[HTML]{34CDF9} 0.330} \\
{\color[HTML]{000000} Different Entity(Non-con)}                  & {\color[HTML]{F56B00} 0.325} & {\color[HTML]{000000} 0.505} & {\color[HTML]{000000} 0.385} & {\color[HTML]{000000} 0.440} & {\color[HTML]{34CDF9} 0.670} & {\color[HTML]{000000} 0.365} & {\color[HTML]{000000} 0.610} \\
{\color[HTML]{000000} Negation(Non-con)}                          & {\color[HTML]{000000} 0.465} & {\color[HTML]{000000} 0.570} & {\color[HTML]{F56B00} 0.380} & {\color[HTML]{000000} 0.600} & {\color[HTML]{34CDF9} 1.310} & {\color[HTML]{000000} 0.565} & {\color[HTML]{000000} 0.685} \\
{\color[HTML]{000000} Average}                                    & 0.183 & 0.290 & 0.251 & 0.238 & 0.376 & 0.216 & 0.388 \\ \bottomrule
\end{tabular}
\caption{The sensitivity of glm4-flash to perturbation texts in the title generation task across seven languages.} 
\label{tab:glm4-flash-title} 
\end{table*}

\begin{table*} 
\small
\centering 
\begin{tabular}{lccccccc}
\toprule
\multicolumn{8}{c}{{\color[HTML]{000000} \textbf{Summary Generation Task}}} \\ \cmidrule(lr){2-8}
{\color[HTML]{000000} \textbf{Perturbations}}   & \multicolumn{1}{l}{\textbf{Bengali}} & \multicolumn{1}{l}{\textbf{French}} & \multicolumn{1}{l}{\textbf{German}} & \multicolumn{1}{l}{\textbf{Hindi}} & \multicolumn{1}{l}{\textbf{Telugu}} & \multicolumn{1}{l}{\textbf{Urdu}} & \multicolumn{1}{l}{\textbf{English}} \\ \midrule
{\color[HTML]{000000} Hypernym(Inf)}            & {\color[HTML]{000000} 0.005} & {\color[HTML]{F56B00} -0.070} & {\color[HTML]{F56B00} -0.070} & {\color[HTML]{000000} -0.025} & {\color[HTML]{000000} -0.030} & {\color[HTML]{000000} -0.045} & {\color[HTML]{34CDF9} 0.040} \\
{\color[HTML]{000000} Sentence Deletion(Inf)}   & {\color[HTML]{34CDF9} 0.145} & {\color[HTML]{F56B00} 0.060} & {\color[HTML]{F56B00} 0.060} & {\color[HTML]{000000} 0.125} & {\color[HTML]{000000} 0.110} & {\color[HTML]{000000} 0.110} & {\color[HTML]{000000} 0.090} \\
{\color[HTML]{000000} Improper Connective(Coh)} & {\color[HTML]{F56B00} 0.065} & {\color[HTML]{000000} 0.100} & {\color[HTML]{000000} 0.100} & {\color[HTML]{000000} 0.100} & {\color[HTML]{000000} 0.075} & {\color[HTML]{000000} 0.075} & {\color[HTML]{34CDF9} 0.155} \\
{\color[HTML]{000000} Sentence Exchange(Coh)}   & {\color[HTML]{000000} 0.065} & {\color[HTML]{000000} 0.060} & {\color[HTML]{000000} 0.060} & {\color[HTML]{000000} 0.085} & {\color[HTML]{000000} 0.060} & {\color[HTML]{F56B00} 0.050} & {\color[HTML]{34CDF9} 0.125} \\
{\color[HTML]{000000} Repetition(Flu)}          & {\color[HTML]{000000} 0.185} & {\color[HTML]{000000} 0.160} & {\color[HTML]{000000} 0.160} & {\color[HTML]{000000} 0.160} & {\color[HTML]{000000} 0.135} & {\color[HTML]{F56B00} 0.125} & {\color[HTML]{34CDF9} 0.315} \\
{\color[HTML]{000000} Passive Voice(Flu)}       & {\color[HTML]{000000} 0.045} & {\color[HTML]{34CDF9} 0.060} & {\color[HTML]{34CDF9} 0.060} & {\color[HTML]{000000} 0.005} & {\color[HTML]{000000} -0.005} & {\color[HTML]{000000} -0.015} & {\color[HTML]{F56B00} -0.030} \\
{\color[HTML]{000000} Average}                  & 0.085 & 0.062 & 0.062 & 0.075 & 0.058 & 0.050 & 0.118 \\ \bottomrule
\end{tabular}
\caption{The sensitivity of glm4-flash to perturbation texts in the summary generation task across seven languages.} 
\label{tab:glm4-flash summary generation task}
\end{table*}

\begin{table*} 
\small
\centering 
\begin{tabular}{lccccccc}
\toprule
\multicolumn{8}{c}{{\color[HTML]{000000} \textbf{Title Generation Task}}} \\ \cmidrule(lr){2-8}
\multicolumn{1}{c}{{\color[HTML]{000000} \textbf{Perturbations}}} & {\color[HTML]{000000} \textbf{Bengali}} & {\color[HTML]{000000} \textbf{French}} & {\color[HTML]{000000} \textbf{German}} & {\color[HTML]{000000} \textbf{Hindi}} & {\color[HTML]{000000} \textbf{Telugu}} & {\color[HTML]{000000} \textbf{Urdu}} & {\color[HTML]{000000} \textbf{English}} \\ \midrule
{\color[HTML]{000000} Uncommon Phrase(Sim)}                       & {\color[HTML]{000000} 0.130} & {\color[HTML]{000000} 0.265} & {\color[HTML]{000000} 0.240} & {\color[HTML]{000000} 0.245} & {\color[HTML]{F56B00} 0.105} & {\color[HTML]{000000} 0.270} & {\color[HTML]{34CDF9} 0.430} \\
{\color[HTML]{000000} Complex Sentence(Sim)}                      & {\color[HTML]{000000} 0.040} & {\color[HTML]{000000} 0.105} & {\color[HTML]{000000} 0.055} & {\color[HTML]{000000} 0.115} & {\color[HTML]{F56B00} 0.020} & {\color[HTML]{000000} 0.120} & {\color[HTML]{34CDF9} 0.220} \\
{\color[HTML]{000000} Complement(Non-hal)}                        & {\color[HTML]{000000} 0.180} & {\color[HTML]{000000} 0.265} & {\color[HTML]{000000} 0.230} & {\color[HTML]{000000} 0.255} & {\color[HTML]{F56B00} 0.160} & {\color[HTML]{000000} 0.260} & {\color[HTML]{34CDF9} 0.495} \\
{\color[HTML]{000000} Continuation(Non-hal)}                      & {\color[HTML]{000000} 0.140} & {\color[HTML]{000000} 0.235} & {\color[HTML]{000000} 0.245} & {\color[HTML]{000000} 0.205} & {\color[HTML]{F56B00} 0.125} & {\color[HTML]{000000} 0.225} & {\color[HTML]{34CDF9} 0.345} \\
{\color[HTML]{000000} Different Entity(Non-con)}                  & {\color[HTML]{000000} 1.430} & {\color[HTML]{000000} 1.530} & {\color[HTML]{000000} 1.410} & {\color[HTML]{000000} 1.495} & {\color[HTML]{F56B00} 1.405} & {\color[HTML]{000000} 1.515} & {\color[HTML]{34CDF9} 1.940} \\
{\color[HTML]{000000} Negation(Non-con)}                          & {\color[HTML]{000000} 1.905} & {\color[HTML]{000000} 1.930} & {\color[HTML]{F56B00} 1.630} & {\color[HTML]{000000} 1.895} & {\color[HTML]{000000} 1.890} & {\color[HTML]{000000} 1.915} & {\color[HTML]{34CDF9} 2.845} \\
{\color[HTML]{000000} Average}                                    & 0.638 & 0.722 & 0.635 & 0.702 & 0.618 & 0.718 & 1.046 \\ \bottomrule
\end{tabular}
\caption{The sensitivity of Qwen2.5-72B to perturbation texts in the title generation task across seven languages.} 
\label{tab:Qwen2.5-72b-title} 
\end{table*}

\begin{table*} 
\small
\centering 
\begin{tabular}{lccccccc}
\toprule
\multicolumn{8}{c}{{\color[HTML]{000000} \textbf{Summary Generation Task}}} \\ \cmidrule(lr){2-8}
\multicolumn{1}{c}{{\color[HTML]{000000} \textbf{Perturbations}}} & {\color[HTML]{000000} \textbf{Bengali}} & {\color[HTML]{000000} \textbf{French}} & {\color[HTML]{000000} \textbf{German}} & {\color[HTML]{000000} \textbf{Hindi}} & {\color[HTML]{000000} \textbf{Telugu}} & {\color[HTML]{000000} \textbf{Urdu}} & {\color[HTML]{000000} \textbf{English}} \\ \midrule
{\color[HTML]{000000} Hypernym(Inf)}                              & {\color[HTML]{000000} 0.010} & {\color[HTML]{34CDF9} 0.065} & {\color[HTML]{000000} 0.060} & {\color[HTML]{000000} -0.010} & {\color[HTML]{000000} 0.015} & {\color[HTML]{F56B00} -0.015} & {\color[HTML]{000000} 0.025} \\
{\color[HTML]{000000} Sentence Deletion(Inf)}                     & {\color[HTML]{F56B00} 0.010} & {\color[HTML]{34CDF9} 0.141} & {\color[HTML]{000000} 0.105} & {\color[HTML]{000000} 0.105} & {\color[HTML]{000000} 0.105} & {\color[HTML]{000000} 0.110} & {\color[HTML]{000000} 0.105} \\
{\color[HTML]{000000} Improper Connective(Coh)}                   & {\color[HTML]{000000} 0.095} & {\color[HTML]{000000} 0.100} & {\color[HTML]{000000} 0.090} & {\color[HTML]{000000} 0.085} & {\color[HTML]{F56B00} 0.010} & {\color[HTML]{000000} 0.095} & {\color[HTML]{34CDF9} 0.105} \\
{\color[HTML]{000000} Sentence Exchange(Coh)}                     & {\color[HTML]{000000} 0.025} & {\color[HTML]{000000} 0.041} & {\color[HTML]{000000} 0.030} & {\color[HTML]{F56B00} 0.015} & {\color[HTML]{34CDF9} 0.135} & {\color[HTML]{000000} 0.020} & {\color[HTML]{000000} 0.030} \\
{\color[HTML]{000000} Repetition(Flu)}                            & {\color[HTML]{000000} 0.210} & {\color[HTML]{000000} 0.242} & {\color[HTML]{000000} 0.230} & {\color[HTML]{F56B00} 0.195} & {\color[HTML]{34CDF9} 0.270} & {\color[HTML]{000000} 0.200} & {\color[HTML]{000000} 0.210} \\
{\color[HTML]{000000} Passive Voice(Flu)}                         & {\color[HTML]{000000} 0.080} & {\color[HTML]{000000} 0.085} & {\color[HTML]{34CDF9} 0.095} & {\color[HTML]{F56B00} 0.075} & {\color[HTML]{000000} 0.100} & {\color[HTML]{000000} 0.080} & {\color[HTML]{000000} 0.085} \\
{\color[HTML]{000000} Average}                                    & 0.072 & 0.112 & 0.102 & 0.078 & 0.106 & 0.082 & 0.093 \\ \bottomrule
\end{tabular}
\caption{The sensitivity of Qwen2.5-72B to perturbation texts in the summary generation task across seven languages.}
\label{tab:Qwen2.5-72B summary generation task}
\end{table*}

\begin{table*} 
\small
\centering 
\begin{tabular}{lccccccc}
\toprule
\multicolumn{8}{c}{{\color[HTML]{000000} \textbf{Title Generation Task}}} \\ \cmidrule(lr){2-8}
\multicolumn{1}{c}{{\color[HTML]{000000} \textbf{Perturbations}}} & {\color[HTML]{000000} \textbf{Bengali}} & {\color[HTML]{000000} \textbf{French}} & {\color[HTML]{000000} \textbf{German}} & {\color[HTML]{000000} \textbf{Hindi}} & {\color[HTML]{000000} \textbf{Telugu}} & {\color[HTML]{000000} \textbf{Urdu}} & {\color[HTML]{000000} \textbf{English}} \\ \midrule
{\color[HTML]{000000} Uncommon Phrase(Sim)}                       & {\color[HTML]{34CDF9} 0.215} & {\color[HTML]{000000} 0.135} & {\color[HTML]{000000} 0.095} & {\color[HTML]{F56B00} 0.063} & {\color[HTML]{000000} 0.085} & {\color[HTML]{000000} 0.090} & {\color[HTML]{000000} 0.170} \\
{\color[HTML]{000000} Complex Sentence(Sim)}                      & {\color[HTML]{F56B00} -0.077} & {\color[HTML]{000000} -0.005} & {\color[HTML]{000000} 0.060} & {\color[HTML]{000000} -0.062} & {\color[HTML]{000000} -0.069} & {\color[HTML]{000000} 0.080} & {\color[HTML]{34CDF9} 0.085} \\
{\color[HTML]{000000} Complement(Non-hal)}                        & {\color[HTML]{F56B00} 0.060} & {\color[HTML]{000000} 0.090} & {\color[HTML]{000000} 0.135} & {\color[HTML]{000000} 0.105} & {\color[HTML]{000000} 0.074} & {\color[HTML]{000000} 0.142} & {\color[HTML]{34CDF9} 0.150} \\
{\color[HTML]{000000} Continuation(Non-hal)}                      & {\color[HTML]{000000} 0.172} & {\color[HTML]{000000} 0.170} & {\color[HTML]{000000} 0.220} & {\color[HTML]{000000} 0.155} & {\color[HTML]{F56B00} 0.145} & {\color[HTML]{000000} 0.255} & {\color[HTML]{34CDF9} 0.240} \\
{\color[HTML]{000000} Different Entity(Non-con)}                  & {\color[HTML]{34CDF9} 0.855} & {\color[HTML]{000000} 0.850} & {\color[HTML]{000000} 0.780} & {\color[HTML]{000000} 0.835} & {\color[HTML]{F56B00} 0.653} & {\color[HTML]{000000} 0.760} & {\color[HTML]{000000} 0.825} \\
{\color[HTML]{000000} Negation(Non-con)}                          & {\color[HTML]{000000} 0.611} & {\color[HTML]{34CDF9} 1.140} & {\color[HTML]{000000} 0.715} & {\color[HTML]{000000} 1.130} & {\color[HTML]{F56B00} 0.533} & {\color[HTML]{000000} 0.690} & {\color[HTML]{000000} 0.715} \\
{\color[HTML]{000000} Average}                                    & 0.306 & 0.397 & 0.334 & 0.371 & 0.237 & 0.336 & 0.364 \\ \bottomrule
\end{tabular}
\caption{The sensitivity of Llama-3.1-70B-Instruct to perturbation texts in the title generation task across seven languages.} 
\label{tab:Llama-3.1-70B-Instruct-title} 
\end{table*}

\begin{table*} 
\small
\centering 
\begin{tabular}{lccccccc}
\toprule
\multicolumn{8}{c}{{\color[HTML]{000000} \textbf{Summary Generation Task}}} \\ \cmidrule(lr){2-8}
\multicolumn{1}{c}{{\color[HTML]{000000} \textbf{Perturbations}}} & {\color[HTML]{000000} \textbf{Bengali}} & {\color[HTML]{000000} \textbf{French}} & {\color[HTML]{000000} \textbf{German}} & {\color[HTML]{000000} \textbf{Hindi}} & {\color[HTML]{000000} \textbf{Telugu}} & {\color[HTML]{000000} \textbf{Urdu}} & {\color[HTML]{000000} \textbf{English}} \\ \midrule
{\color[HTML]{000000} Hypernym(Inf)}                              & {\color[HTML]{000000} 0.030} & {\color[HTML]{000000} 0.125} & {\color[HTML]{34CDF9} 0.170} & {\color[HTML]{000000} -0.090} & {\color[HTML]{F56B00} -0.120} & {\color[HTML]{000000} -0.090} & {\color[HTML]{000000} 0.055} \\
{\color[HTML]{000000} Sentence Deletion(Inf)}                     & {\color[HTML]{F56B00} 0.070} & {\color[HTML]{34CDF9} 0.195} & {\color[HTML]{000000} 0.130} & {\color[HTML]{000000} 0.145} & {\color[HTML]{000000} 0.140} & {\color[HTML]{000000} 0.140} & {\color[HTML]{000000} 0.185} \\
{\color[HTML]{000000} Improper Connective(Coh)}                   & {\color[HTML]{000000} 0.060} & {\color[HTML]{000000} 0.065} & {\color[HTML]{000000} 0.035} & {\color[HTML]{F56B00} 0.030} & {\color[HTML]{000000} 0.105} & {\color[HTML]{F56B00} 0.030} & {\color[HTML]{34CDF9} 0.240} \\
{\color[HTML]{000000} Sentence Exchange(Coh)}                     & {\color[HTML]{000000} 0.130} & {\color[HTML]{000000} 0.190} & {\color[HTML]{000000} 0.160} & {\color[HTML]{000000} 0.090} & {\color[HTML]{000000} 0.150} & {\color[HTML]{F56B00} 0.085} & {\color[HTML]{34CDF9} 0.210} \\
{\color[HTML]{000000} Repetition(Flu)}                            & {\color[HTML]{000000} 0.080} & {\color[HTML]{000000} 0.095} & {\color[HTML]{F56B00} 0.045} & {\color[HTML]{000000} 0.075} & {\color[HTML]{000000} 0.085} & {\color[HTML]{000000} 0.070} & {\color[HTML]{34CDF9} 0.330} \\
{\color[HTML]{000000} Passive Voice(Flu)}                         & {\color[HTML]{000000} 0.030} & {\color[HTML]{000000} 0.045} & {\color[HTML]{F56B00} -0.015} & {\color[HTML]{000000} 0.110} & {\color[HTML]{34CDF9} 0.165} & {\color[HTML]{000000} 0.100} & {\color[HTML]{000000} 0.150} \\
{\color[HTML]{000000} Average}                                    & 0.067 & 0.119 & 0.088 & 0.060 & 0.088 & 0.056 & 0.195 \\ \bottomrule
\end{tabular}
\caption{The sensitivity of Llama-3.1-70B-Instruct to perturbation texts in the summary generation task across seven languages.}
\label{tab:Llama-3.1-70B-Instruct summary generation task}
\end{table*}

\begin{table*} 
\small
\centering 
\begin{tabular}{lccccccc}
\toprule
\multicolumn{8}{c}{{\color[HTML]{000000} \textbf{Title Generation Task}}} \\ \cmidrule(lr){2-8}
\multicolumn{1}{c}{{\color[HTML]{000000} \textbf{Perturbations}}} & {\color[HTML]{000000} \textbf{Bengali}} & {\color[HTML]{000000} \textbf{French}} & {\color[HTML]{000000} \textbf{German}} & {\color[HTML]{000000} \textbf{Hindi}} & {\color[HTML]{000000} \textbf{Telugu}} & {\color[HTML]{000000} \textbf{Urdu}} & {\color[HTML]{000000} \textbf{English}} \\ \midrule
{\color[HTML]{000000} Uncommon Phrase(Sim)}                       & {\color[HTML]{000000} -0.045} & {\color[HTML]{000000} 0.120} & {\color[HTML]{000000} 0.124} & {\color[HTML]{000000} 0.103} & {\color[HTML]{F56B00} -0.155} & {\color[HTML]{000000} 0.005} & {\color[HTML]{34CDF9} 0.235} \\
{\color[HTML]{000000} Complex Sentence(Sim)}                      & {\color[HTML]{000000} -0.135} & {\color[HTML]{000000} 0.132} & {\color[HTML]{000000} 0.122} & {\color[HTML]{000000} -0.156} & {\color[HTML]{F56B00} -0.240} & {\color[HTML]{000000} -0.145} & {\color[HTML]{34CDF9} 0.175} \\
{\color[HTML]{000000} Complement(Non-hal)}                        & {\color[HTML]{000000} 0.180} & {\color[HTML]{000000} 0.401} & {\color[HTML]{000000} 0.400} & {\color[HTML]{000000} 0.163} & {\color[HTML]{F56B00} 0.085} & {\color[HTML]{34CDF9} 0.525} & {\color[HTML]{34CDF9} 0.525} \\
{\color[HTML]{000000} Continuation(Non-hal)}                      & {\color[HTML]{000000} 0.140} & {\color[HTML]{000000} 0.302} & {\color[HTML]{000000} 0.296} & {\color[HTML]{000000} 0.139} & {\color[HTML]{F56B00} 0.050} & {\color[HTML]{34CDF9} 0.490} & {\color[HTML]{000000} 0.375} \\
{\color[HTML]{000000} Different Entity(Non-con)}                  & {\color[HTML]{000000} 1.395} & {\color[HTML]{000000} 1.961} & {\color[HTML]{000000} 1.542} & {\color[HTML]{F56B00} 0.122} & {\color[HTML]{000000} 1.405} & {\color[HTML]{000000} 1.645} & {\color[HTML]{34CDF9} 2.005} \\
{\color[HTML]{000000} Negation(Non-con)}                          & {\color[HTML]{000000} 1.870} & {\color[HTML]{000000} 1.540} & {\color[HTML]{000000} 1.496} & {\color[HTML]{F56B00} 0.130} & {\color[HTML]{000000} 1.290} & {\color[HTML]{000000} 2.045} & {\color[HTML]{000000} 1.910} \\
{\color[HTML]{000000} Average}                                    & 0.568 & 0.743 & 0.663 & 0.084 & 0.406 & 0.761 & 0.871 \\ \bottomrule
\end{tabular}
\caption{The sensitivity of Qwen-turbo to perturbation texts in the title generation task across seven languages.} 
\label{tab: Qwen-turbo-title} 
\end{table*}

\begin{table*} 
\small
\centering 
\begin{tabular}{lccccccc}
\toprule
\multicolumn{8}{c}{{\color[HTML]{000000} \textbf{Summary Generation Task}}} \\ \cmidrule(lr){2-8}
\multicolumn{1}{c}{{\color[HTML]{000000} \textbf{Perturbations}}} & {\color[HTML]{000000} \textbf{Bengali}} & {\color[HTML]{000000} \textbf{French}} & {\color[HTML]{000000} \textbf{German}} & {\color[HTML]{000000} \textbf{Hindi}} & {\color[HTML]{000000} \textbf{Telugu}} & {\color[HTML]{000000} \textbf{Urdu}} & {\color[HTML]{000000} \textbf{English}} \\ \midrule
{\color[HTML]{000000} Hypernym(Inf)}                              & {\color[HTML]{000000} -0.005} & {\color[HTML]{000000} 0.095} & {\color[HTML]{000000} 0.110} & {\color[HTML]{000000} -0.015} & {\color[HTML]{000000} 0.005} & {\color[HTML]{F56B00} -0.035} & {\color[HTML]{000000} 0.015} \\
{\color[HTML]{000000} Sentence Deletion(Inf)}                     & {\color[HTML]{F56B00} 0.090} & {\color[HTML]{000000} 0.125} & {\color[HTML]{000000} 0.140} & {\color[HTML]{000000} 0.100} & {\color[HTML]{000000} 0.095} & {\color[HTML]{000000} 0.155} & {\color[HTML]{34CDF9} 0.195} \\
{\color[HTML]{000000} Improper Connective(Coh)}                   & {\color[HTML]{000000} 0.040} & {\color[HTML]{000000} 0.245} & {\color[HTML]{34CDF9} 0.255} & {\color[HTML]{000000} 0.005} & {\color[HTML]{000000} 0.095} & {\color[HTML]{F56B00} -0.020} & {\color[HTML]{34CDF9} 0.255} \\
{\color[HTML]{000000} Sentence Exchange(Coh)}                     & {\color[HTML]{000000} 0.110} & {\color[HTML]{34CDF9} 0.300} & {\color[HTML]{000000} 0.205} & {\color[HTML]{000000} 0.065} & {\color[HTML]{000000} 0.020} & {\color[HTML]{F56B00} -0.110} & {\color[HTML]{000000} 0.220} \\
{\color[HTML]{000000} Repetition(Flu)}                            & {\color[HTML]{F56B00} 0.030} & {\color[HTML]{000000} 0.120} & {\color[HTML]{000000} 0.140} & {\color[HTML]{000000} 0.055} & {\color[HTML]{34CDF9} 0.365} & {\color[HTML]{000000} 0.045} & {\color[HTML]{000000} 0.105} \\
{\color[HTML]{000000} Passive Voice(Flu)}                         & {\color[HTML]{F56B00} -0.100} & {\color[HTML]{000000} 0.255} & {\color[HTML]{000000} 0.215} & {\color[HTML]{000000} -0.061} & {\color[HTML]{000000} 0.095} & {\color[HTML]{000000} 0.070} & {\color[HTML]{34CDF9} 0.420} \\
{\color[HTML]{000000} Average}                                    & 0.178 & 0.190 & 0.178 & 0.025 & 0.113 & 0.018 & 0.202 \\ \bottomrule
\end{tabular}
\caption{The sensitivity of Qwen-turbo to perturbation texts in the summary generation task across seven languages.}
\label{tab:Qwen-turbo summary generation task}
\end{table*}

\section{The multilingual evaluation capabilities of LLMs with reference answers}
\label{sec:with and without reference answers}

The experimental results of The multilingual evaluation capabilities of LLMs with reference answers are shown in \autoref{tab:with reference answers}.

\begin{table*}[htbp] 
    \centering 
    \includegraphics{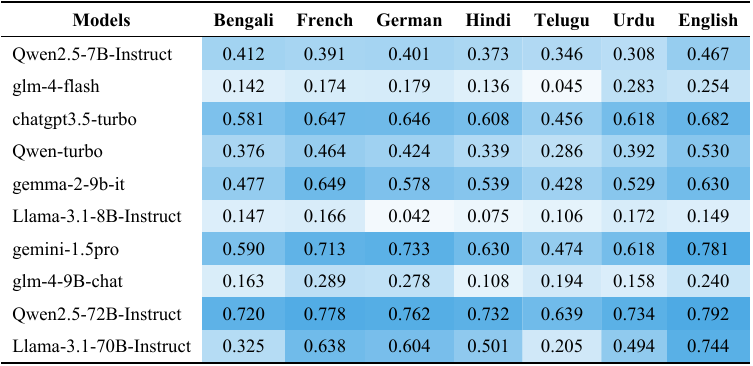} 
    \caption{The multilingual evaluation capabilities of LLMs with reference answers.}
    \label{tab:with reference answers}
\end{table*}

\section{The language of the publicly available training data for LLMs.}
\label{sec:The language of the publicly available training data for LLMs}

The languages of the publicly available training data for the ten LLMs utilized in this study are outlined in \autoref{tab:The languages of the publicly available training data}. Notably, only Llama-3.1-8B-Instruct, Gemini-1.5Pro, and Llama-3.1-70B-Instruct have fully disclosed the languages included in their training data, while the remaining models offer only broad or imprecise language listings.

\begin{table*}[]
\small
\centering 
\begin{tabular}{cl}
\toprule
{\color[HTML]{000000} \textbf{Models}}        & \multicolumn{1}{c}{{\color[HTML]{000000} \textbf{Language}}}                                                                                                                                                                                                                                                                                                                                                                                                                   \\ \midrule
{\color[HTML]{000000} glm-4-flash}            & {\color[HTML]{000000} \begin{tabular}[c]{@{}l@{}}Twenty-six languages, including Chinese, English, \\ Japanese, Korean, German, and others.\end{tabular}}                                                                                                                                                                                                                                                                                                                      \\ \cline{2-2} 
{\color[HTML]{000000} chatgpt3.5-turbo}       & {\color[HTML]{000000} The language of the training data has not been disclosed.}                                                                                                                                                                                                                                                                                                                                                                                                          \\ \cline{2-2} 
{\color[HTML]{000000} Qwen-turbo}             & {\color[HTML]{000000} It supports multiple languages, including Chinese and English.}                                                                                                                                                                                                                                                                                                                                                                                          \\ \cline{2-2} 
{\color[HTML]{000000} gemma-2-9b-it}          & {\color[HTML]{000000} Primarily for English content.}                                                                                                                                                                                                                                                                                                                                                                                                                          \\ \cline{2-2} 
{\color[HTML]{009901} gemini-1.5pro}          & \begin{tabular}[c]{@{}l@{}}Arabic, Bengali, Bulgarian, Simplified Chinese, Traditional Chinese, \\ Croatian, Czech, Danish, Dutch, English, Estonian, Finnish, French, \\ German, Greek, Hebrew, Hindi, Hungarian, Indonesian, Italian, Japanese, \\ Korean, Latvian, Lithuanian, Norwegian, Polish, Portuguese, Romanian, \\ Russian, Serbian, Slovak, Slovenian, Spanish, Swahili, Swedish, Thai, \\ Turkish, Ukrainian, and Vietnamese, totaling 38 languages.\end{tabular} \\ \cline{2-2} 
{\color[HTML]{000000} glm-4-9B-chat}          & {\color[HTML]{000000} It supports 26 languages, including Japanese, Korean, and German.}                                                                                                                                                                                                                                                                                                                                                                                       \\ \cline{2-2} 
{\color[HTML]{000000} Qwen2.5-72B}            & {\color[HTML]{000000} \begin{tabular}[c]{@{}l@{}}It supports over 29 languages, including Chinese, English, French, \\ Spanish, Portuguese, German, Italian, Russian, Japanese, Korean, \\ Vietnamese, Thai, Arabic, and more.\end{tabular}}                                                                                                                                                                                                                                   \\ \cline{2-2} 
{\color[HTML]{009901} Llama-3.1-70B-Instruct} & {\color[HTML]{000000} \begin{tabular}[c]{@{}l@{}}In addition to supporting English, it also supports seven languages: \\ French, German, Hindi, Italian, Portuguese, Spanish, and Thai.\end{tabular}}                                                                                                                                                                                                                                                                          \\ \bottomrule
\end{tabular}
\caption{The languages of the publicly available training data for the ten LLMs discussed in this paper are presented, with those highlighted in green representing the models that have fully disclosed their training languages.}
\label{tab:The languages of the publicly available training data}

\end{table*}

\end{document}